\documentclass[lettersize,journal]{IEEEtran}
\usepackage{amsmath,amsfonts}
\usepackage{algorithmic}
\usepackage{algorithm}
\usepackage{array}
\usepackage[caption=false,font=normalsize,labelfont=sf,textfont=sf]{subfig}
\usepackage{textcomp}
\usepackage{stfloats}
\usepackage{url}
\usepackage{verbatim}
\usepackage{booktabs}
\usepackage{arydshln}
\usepackage[dvipdfmx]{graphicx}
\usepackage{cite}
\usepackage{siunitx}
\usepackage{multirow}
\usepackage{pifont}
\newcommand{\figref}[1]{Fig.~\ref{#1}}
\newcommand{\tabref}[1]{Table~\ref{#1}}

\begin{document}

\title{Channel-Free Human Activity Recognition via Inductive-Bias-Aware Fusion Design for Heterogeneous IoT Sensor Environments}

\author{Tatsuhito Hasegawa,~\IEEEmembership{Member,~IEEE,}
\thanks{Manuscript received April 19, 2026; revised August xx, 202x. This work was supported in part by the Japan Society for the Promotion of Science (JSPS) KAKENHI Grant-in-Aid for Scientific Research (B) and (C) under Grant 26K02879 and 23K11164, and in part by the Several Competitive Funds within the University of Fukui.}
\thanks{Tatsuhito Hasegawa is with the Graduate School of Engineering, University of Fukui, Bunkyo, Fukui, 9108507, Fukui, Japan (e-mail:t-hase@u-fukui.ac.jp)}}

\markboth{Journal of \LaTeX\ Class Files,~Vol.~14, No.~8, August~2021}%
{Shell \MakeLowercase{\textit{et al.}}: A Sample Article Using IEEEtran.cls for IEEE Journals}


\maketitle

\begin{abstract}
Human activity recognition (HAR) in Internet of Things (IoT) environments must cope with heterogeneous sensor settings that vary across datasets, devices, body locations, sensing modalities, and channel compositions. This heterogeneity makes conventional channel-fixed models difficult to reuse across sensing environments because their input representations are tightly coupled to predefined channel structures. To address this problem, we investigate strict channel-free HAR, in which a single shared model performs inference without assuming a fixed number, order, or semantic arrangement of input channels, and without relying on sensor-specific input layers or dataset-specific channel templates. We argue that fusion design is the central issue in this setting. Accordingly, we propose a channel-free HAR framework that combines channel-wise encoding with a shared encoder, metadata-conditioned late fusion via conditional batch normalization, and joint optimization of channel-level and fused predictions through a combination loss. The proposed model processes each channel independently to handle varying channel configurations, while sensor metadata such as body location, modality, and axis help recover structural information that channel-independent processing alone cannot retain. In addition, the joint loss encourages both the discriminability of individual channels and the consistency of the final fused prediction. Experiments on PAMAP2, together with robustness analysis on six HAR datasets, ablation studies, sensitivity analysis, efficiency evaluation, and cross-dataset transfer learning, demonstrate three main findings: 1) channel-free late fusion achieves strong robustness under channel perturbations; 2) metadata conditioning consistently improves the channel-free model while remaining tolerant to mild signal--metadata inconsistency; and 3) the resulting framework remains competitive with conventional channel-fixed architectures while enabling transfer across heterogeneous sensor configurations without input-layer redesign. These results suggest that channel-free modeling designed from inductive-bias considerations is a practical step toward scalable and transferable HAR in real-world IoT environments.
\end{abstract}


\section{Introduction}
\IEEEPARstart{H}{uman} activity recognition (HAR) aims to infer a wearer’s activities from diverse sensor signals and is a key technology for Internet of Things (IoT) applications such as health monitoring, assisted living, sports analytics, and context-aware services. As summarized in \tabref{table:datasets}, existing sensor-based HAR studies cover highly diverse input settings. Many benchmark datasets focus on daily activity recognition using only a tri-axial accelerometer \cite{Kwapisz2011, Kawaguchi2011, Daniela2017}, whereas others additionally employ gyroscopes and magnetometers \cite{Anguita2013-qb, Shoaib2016}, or place sensors on multiple body locations such as the hand, chest, ankle, and waist \cite{Stisen2015-ai, Banos2014-uc, Altun2010-kp, Reiss2012a}. In general, increasing the number and variety of sensors enables richer observation of real-world human behavior, but it also makes data processing substantially more complex. In this sense, sensor-based HAR is inherently a multimodal learning problem in which the number, type, and semantic meaning of channels vary markedly across datasets and deployment environments.

\begin{table*}[t]
    \caption{Benchmark datasets of sensor-based HAR. The ``\# of ch.'' column lists the raw channel count reported by each dataset; as described in Section~\ref{sec:setup}, PAMAP2 is used with 40 channels in our experiments because the orientation measurements are excluded due to known recording issues.}
    \label{table:datasets}
    \hbox to\hsize{\hfil
    \begin{tabular}{cccccccc} \hline \hline
    Dataset & Cite & Sensors & Location & \# of ch. & \# of act. & \# of subj. & Freq. [Hz] \\ \hline
    WISDM & \cite{Kwapisz2011} & A & Pocket & 3 & 6 & 29 & 20 \\
    HASC & \cite{Kawaguchi2011} & A & - & 3 & 6 & 116 & 10--100$^{\dagger}$ \\
    UniMiB SHAR & \cite{Daniela2017} & A & Pocket & 3 & 17 & 30 & $\approx$50 \\
    UCI-HAR & \cite{Anguita2013-qb} & A,G & Waist & 9 & 6 & 30 & 50 \\
    CHAD & \cite{Shoaib2016} & A,G,M & Pocket, Wrist & 12 & 13 & 10 & 50 \\
    HHAR & \cite{Stisen2015-ai} & A,G & Waist, Hand & 12 & 6 & 9 & 50--200$^{\ddagger}$ \\
    MHEALTH & \cite{Banos2014-uc} & A,G,M,H & Hand, Chest, Ankle & 23 & 12 & 10 & 50 \\
    DSADS & \cite{Altun2010-kp} & A,G,M & 5 positions & 45 & 19 & 8 & 25 \\
    PAMAP2 & \cite{Reiss2012a} & A,G,M,H,T & Hand, Chest, Ankle & 52 & 18 & 9 & 100$^{\S}$ \\ \hline
    \multicolumn{8}{l}{A: accelerometer, G: gyroscope, M: magnetic, H: heart rate, and T: temperature.} \\
    \multicolumn{8}{l}{$^{\dagger}$ HASC corpus stores recordings with varying sampling rates; commonly reported as 10--100 Hz.} \\
    \multicolumn{8}{l}{$^{\ddagger}$ HHAR is device-dependent; common smartphone rates span 50, 100, 150, and 200 Hz.} \\
    \multicolumn{8}{l}{$^{\S}$ PAMAP2 IMUs are sampled at 100 Hz, while the heart-rate monitor is approximately 9 Hz.}
    \end{tabular}\hfil}
\end{table*}

Despite this diversity, most existing HAR algorithms are implemented using machine learning or deep learning under a \textbf{channel-fixed} assumption \cite{Kwapisz2011, Dong2019}. Typical approaches either extract features from individual channels separately \cite{Kwapisz2011} or jointly convolve all channels from the first layer \cite{Dong2019}. In both cases, the model is built for a predefined input structure, that is, a fixed number, order, and semantic meaning of channels. For example, a model trained for only three channels of right-hand acceleration is not directly applicable to a different configuration that additionally includes three channels of left-hand gyroscope data. Consequently, different sensor configurations usually require different models, even when the target task itself remains the same. This design is workable in closed benchmark settings, but it fundamentally limits scalability across heterogeneous IoT sensing environments.

In contrast, foundation models have become the dominant paradigm in computer vision and natural language processing \cite{bommasani2022opportunities}. A foundation model is a large-scale pretrained model that can be transferred to a wide range of downstream tasks. Representative examples include Grounding DINO for object detection \cite{GDINO2025}, SAM for segmentation \cite{SAM2023}, and GPT for natural language processing \cite{GPT2018}. Such models are realized by training high-capacity architectures on extremely large and diverse datasets, with or without dense task-specific annotations. Their success suggests that, rather than designing a separate model for each individual benchmark, a more scalable direction is to build a single model family that can absorb heterogeneous data distributions and transfer across tasks and domains.

To realize foundation models for HAR, at least two requirements must be satisfied: (1) large-scale training data and (2) a model architecture that can uniformly process sensor time series with different channel configurations. Regarding the first requirement, as shown in \tabref{table:datasets}, many HAR datasets are available, but each individual dataset is relatively limited in scale compared with the data regimes commonly used for foundation models. To alleviate this issue, recent studies have begun to explore cross-dataset utilization, such as domain generalization \cite{Qin2020, Hong2024} and data-mixtured pretraining \cite{Zhao2022_DRP, ban2025}. These directions suggest that heterogeneous HAR datasets can be jointly exploited to improve generalization. However, the second requirement remains insufficiently addressed: existing HAR models are still largely tied to dataset-specific channel structures, making it difficult to learn from multiple datasets in a unified and scalable manner.

From this perspective, we argue that \textbf{channel-free} design is a key requirement for future HAR foundation models. Here, ``channel-free'' means that a model can accept sensor inputs with different numbers and compositions of channels without assuming a fixed channel template in advance. This property is essential when training on heterogeneous IoT sensor data collected from different devices, sensing modalities, and body locations. At the same time, achieving high accuracy under a channel-free constraint is nontrivial. If channel-specific assumptions are removed naively, the model may lose structural biases that are implicitly exploited by conventional channel-fixed architectures. Therefore, the core challenge is not merely to make HAR models flexible to variable channels, but to design a channel-free architecture that preserves or even improves recognition performance.

The essential difficulty of channel-free HAR lies in treating the input not as a fixed-length ordered vector, but as a variable-size and unordered set of channels. More specifically, the input is represented as a set of sensor time series $\{\mathcal{X}_i\}_{i=1}^{c}$, where $\mathcal{X}_i$ denotes the time series of channel $i$. Such a formulation is naturally related to permutation-invariant modeling, as typified by DeepSets \cite{NIPS2017_f22e4747}, because the prediction should not depend on an arbitrary ordering of channels. However, this constraint does not uniquely determine the model architecture. There remain multiple design choices regarding when and how to integrate channel information, such as fusing channels immediately at the input stage, mixing them in intermediate representations, or encoding them independently before integration. These alternatives impose different inductive biases on cross-channel interaction and representation learning.

In this paper, we address this design problem by systematically reconsidering fusion strategies from the viewpoint of channel-free HAR. Our main claim is that the inductive bias of fusion architecture is critical to achieving both channel flexibility and high recognition accuracy. Based on this perspective, we develop a channel-free HAR model that combines shared encoder-based channel-wise feature extraction, metadata injection using sensor location, sensor modality, and axis information, and a fusion design tailored to the channel-free setting. The proposed architecture is intended not only to tolerate heterogeneous sensor configurations but also to provide a practical architectural basis for future data-mixtured pretraining across multiple HAR datasets. Extensive experiments further show that an appropriately designed channel-free model can achieve robust performance across heterogeneous settings and even surpass conventional channel-fixed architectures.

The main contributions of this article are summarized as follows:
\begin{itemize}
    \item We formalize \textbf{strict channel-free HAR} as a new problem setting for future HAR foundation models, where a single shared model must process heterogeneous sensor inputs with different numbers and compositions of channels without relying on a predefined channel template or any input-dependent preprocessing module. This setting is strictly stronger than the ``variable-format'' formulations considered in prior work including the authors' previous study~\cite{Hasegawa2024}, which still required per-input linear transformations as a STEM.
    \item We analyze the inductive biases of fusion design choices under the strict channel-free constraint and derive a model design that combines channel-wise encoding with a shared encoder (no input-dependent STEM), explicit metadata conditioning via conditional batch normalization, and a combination loss that jointly supervises fused and channel-wise predictions.
    \item Through extensive experiments, including benchmark evaluation on PAMAP2, robustness comparison across six datasets, five backbone architectures, ablation studies, perturbation-intensity sweeps, sensitivity analyses, efficiency measurement, and cross-dataset transfer learning, we show that the proposed channel-free design achieves strong robustness under channel perturbations and can even outperform conventional channel-fixed models.
\end{itemize}

\section{Related Work}
\subsection{Sensor-Based HAR}
Sensor-based HAR has been applied not only to the recognition of daily activities such as walking and running using wearable sensors including tri-axial accelerometers \cite{Kwapisz2011, Kawaguchi2011, Daniela2017}, but also to a wide range of IoT applications such as nursing activity recognition \cite{Inoue2015} and work recording in factory environments \cite{Maekawa2016}. Research in this field can be broadly divided into sensing design, which concerns what sensors are used and how they are worn for data acquisition, and recognition model design, which concerns how the acquired signals are processed for recognition. This study focuses on the latter, namely, how heterogeneous sensor signals can be processed in a unified manner by an appropriate model architecture.

Conventional HAR has long been realized using machine learning. In classical approaches, hand-crafted features such as waveform amplitude and frequency characteristics are extracted from each sensor signal, and activities are recognized using classifiers such as Random Forest \cite{Kwapisz2011, Shoaib2016, Robert2019}. In contrast, recent studies have increasingly adopted deep learning models, particularly convolutional neural networks (CNNs), which jointly learn feature extraction and classification \cite{Dong2019, Tuncer2020, Ronald2021, Zhao2022, Miah2024, YANG2025108314, Wan2020-iz, Dua2021-dz}. In general, when sufficient labeled training data are available, deep learning-based methods often achieve higher performance than methods based on manually designed features.

However, many of these methods are designed under the assumption that the number of input channels and the sensor configuration are fixed in advance. In other words, each target dataset prespecifies which sensors are used and how many channels are input, and the model is optimized for that fixed configuration. Although this assumption is effective within a single dataset, it severely limits model reusability and scalability in IoT environments, where sensor modality, body location, and channel composition may vary. Therefore, despite the recent advances in deep learning-based HAR, a fundamental challenge remains in handling heterogeneous sensor configurations with a single model.

\subsection{Cross-Dataset Learning for HAR}

With the progress of deep learning-based HAR, research has also advanced toward enlarging training data and learning more generalizable feature representations. The most standard direction is data augmentation within a single dataset. Representative examples include geometric transformations applied to training signals \cite{RASHID2019100944}, synthetic data generation \cite{Hasegawa2021}, and the use of generative models \cite{Li2020}. More recently, large-scale pretraining for general representation learning has also been reported using 59.7 million hours of wearable sensor data held by Google \cite{zhang2025sensorlm}.

In addition, increasing attention has been paid to studies that leverage multiple datasets in a cross-dataset manner. Typical directions include transfer learning, which transfers knowledge pretrained on one dataset to another \cite{Varshney2022}; domain adaptation, which reduces the representation gap between source and target domains \cite{Zhao2021}; domain generalization, which learns feature representations robust to domain shift \cite{Qin2020, Hong2024}; and Data Mixtured Pretraining, which pretrains a model using a mixture of multiple datasets \cite{Zhao2022_DRP, ban2025, Miao2024, Qiu2025}. These studies represent an important direction toward HAR foundation models, as they aim to improve generalization by integrating knowledge from multiple datasets under the common constraint that each individual HAR dataset is relatively small.

Nevertheless, many existing studies assume that the input shape is essentially common across datasets, or at least that dataset-specific input transformations are introduced for each dataset \cite{Zhao2022_DRP, ban2025}. In fact, even studies that explicitly consider differences in input shape typically prepare different encoders or linear layers for each sensor or dataset in advance, rather than aiming to let a single shared model directly accept heterogeneous channel configurations as they are \cite{Qiu2025}. Therefore, although existing cross-dataset learning studies address the problem of leveraging multiple datasets, they do not sufficiently address the channel-free requirement of handling different channel configurations in a unified manner with a single model.

\subsection{Multimodal Learning and Fusion Design}
\begin{figure}[b]
    \centering
    \includegraphics[width=1.\linewidth]{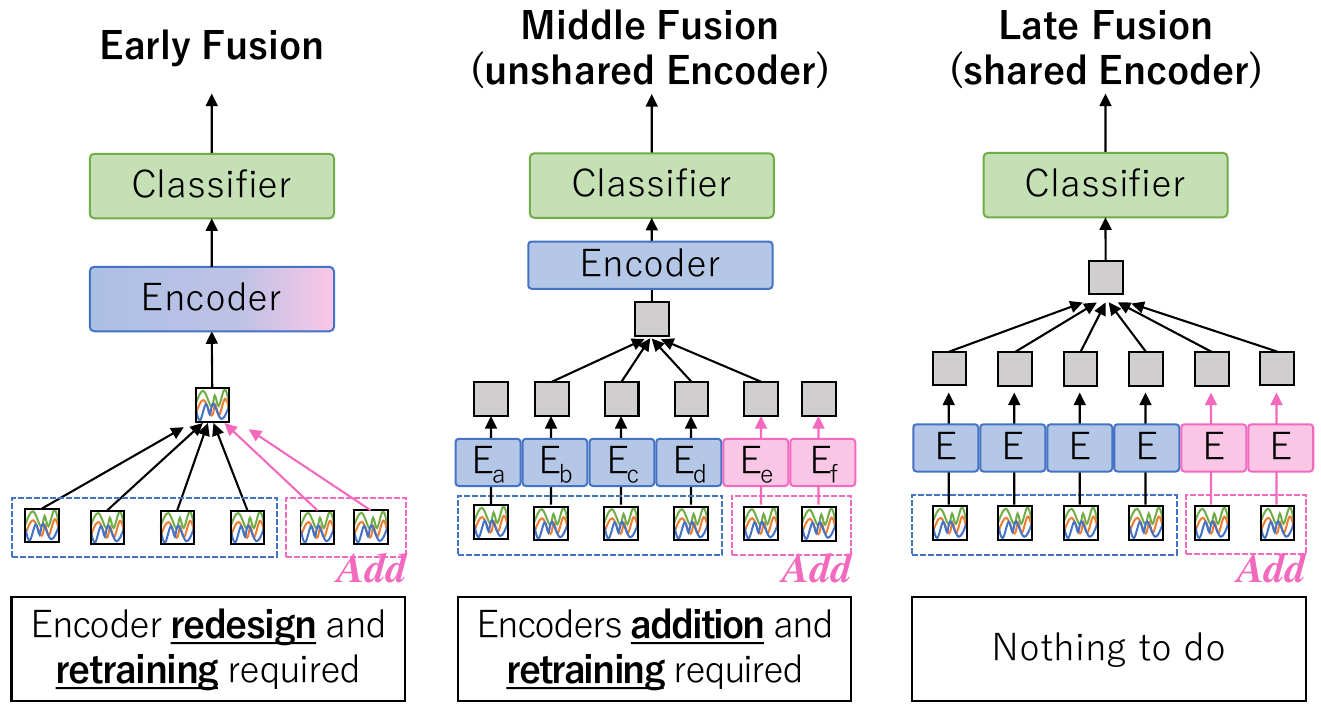}
    \caption{Comparison of EF, MF, and LF for channel-free HAR. The three strategies differ in the stage at which channel information is integrated, whether channel encoders are shared, and how naturally they accommodate newly added channels.}
    \label{fig:fusion}
\end{figure}
A framework that learns by integrating multiple modalities or multiple sensors is broadly referred to as multimodal learning. In multimodal recognition, fusion strategies are commonly categorized according to the stage at which information is integrated, such as Early Fusion (EF), which fuses inputs before or at the early stage of feature extraction \cite{mo2024deepavfusion}, and Late Fusion (LF), which integrates modality-specific features after separate encoding \cite{Li2024AV}. These ideas can be naturally extended to sensor-based HAR, where fusion design can likewise be organized into EF, Middle Fusion (MF), and LF, as illustrated in \figref{fig:fusion}. EF combines multiple sensors or channels before the encoder, which allows strong cross-channel interactions to be learned from the earliest stage, but also makes the encoder tightly coupled to a fixed input structure; as a result, adding a new channel often requires redesigning and retraining the encoder itself. MF introduces channel interactions in intermediate layers, and is therefore more flexible than EF in principle, but when separate encoders are used for different sensors or channels, additional dedicated encoders must still be introduced and trained as the input configuration changes. In contrast, LF first encodes each sensor or channel independently and then aggregates the resulting features, making it structurally more extensible; in particular, when a shared encoder is adopted, newly added channels can be processed without modifying the backbone itself, and only the aggregation step needs to accommodate the changed channel set. Thus, beyond the difference in fusion timing, EF, MF, and LF also differ in how naturally they support variable and evolving sensor configurations, which is a central issue in channel-free HAR.

Many existing deep learning-based HAR methods assume EF \cite{Dong2019, Tuncer2020, Ronald2021, Zhao2022, Miah2024, YANG2025108314, Wan2020-iz, Dua2021-dz}. EF can strongly learn cross-channel interactions from an early stage, but it also tends to implicitly assume that the number, order, and semantic meaning of the input channels are fixed. As a result, it is disadvantageous when heterogeneous sensor configurations must be handled flexibly. In contrast, LF first processes each sensor or each channel more independently and then integrates the resulting features, making it more naturally compatible with variable sensor configurations. However, simply introducing LF may also lose the structural information across channels that EF implicitly exploits, which may lead to insufficient recognition performance. Therefore, fusion design in HAR should be understood not merely as a choice of integration method, but as a choice of inductive bias that determines how to balance configurational flexibility and recognition accuracy.

A closer look at LF-style HAR methods allows a finer-grained categorization, as summarized in \tabref{table:lfmethods}. The Input column indicates whether the unit of input to the encoder is a sensor or a channel. In sensor-level input, each sensor is treated as a fixed channel set, such as a tri-axial accelerometer; therefore the input structure is fixed for each sensor. In contrast, channel-level input decomposes each sensor into individual channels, enabling a more flexible representation for different sensor modalities and channel configurations. The Encoder column indicates whether the encoder is shared or independently prepared for each sensor or channel. When independent encoders are used, separate training is required for different sensor modalities or channel counts, which reduces scalability as a shared model. The Fusion column indicates the integration operation before the classifier. However, many fusion methods, including convolution, concatenation, and stacking, assume fixed-length input and therefore cannot directly handle a varying number of sensors or channels. Even attention-based methods often provide only partial support for variable-length input. Thus, even existing LF-style methods often do not satisfy channel-freedom in the strict sense.

\subsection{Position of This Work}

There exist studies that learn across datasets with different input shapes \cite{Miao2024, Qiu2025}. However, these approaches typically rely on dataset-specific linear transformations or dedicated modules before or after the encoder, and are not intended to allow a single shared model to accept inputs from unseen sensor configurations or unseen datasets without a fixed channel template. The authors' previous work~\cite{Hasegawa2024} also explored variable input formats and provided an important starting point for handling heterogeneous sensor configurations through input-combining, parameter-sharing, and output-combining strategies together with dynamic grouping. The present work extends that line of research in four main respects:

\begin{table}[t] 
    \caption{Comparison of LF-style multimodal HAR methods.}
    \label{table:lfmethods}
    \hbox to\hsize{\hfil
    \begin{tabular}{ccccc} \hline \hline
    Cite &	Input &	Encoder &	Fusion &	Ch-Free \\ \hline
    \cite{Baloch2022-sl} &	Sensor &	Indep. &	Convolution &	No  \\ 
    \cite{Yang2018-co} &	Sensor &	Shared &	Concatenate &	No  \\ 
    \cite{Li2025} &	Sensor &	Shared &	Concatenate &	No  \\ 
    \cite{Zhao2023-kt} &	Sensor &	Shared &	Attention &	Partial  \\ 
    \cite{Noori2020} &	Sensor &	Indep. &	Stacking &	No  \\ 
    \cite{Gholamrezaii2021-da} &	Channel &	Indep. &	Concatenate &	No  \\ 
    \cite{Kasnesis2019} &	Channel &	Shared &	Convolution &	No  \\ 
    \cite{Munzner2017-uw} &	Both &	Both &	Concatenate &	Partial  \\ 
    \cite{Hasegawa2024} &	Both &	Both &	Sum/Concatenate &	Partial  \\ \hline
    Ours &	Channel &	Shared &	Mean + Aux CE &	\textbf{Yes}  \\ \hline
    \end{tabular}\hfil}
\end{table} 
\begin{enumerate}
    \item[(i)] \textbf{Channel-freedom.} 
    In \cite{Hasegawa2024}, a per-input linear STEM (Conv1D($n\!\to\!3$) + BatchNorm) was introduced so that downstream encoders could operate on a fixed three-channel tensor. While this design enabled partial flexibility with respect to input format, the STEM itself still depended on the input shape. In the present work, we further simplify the framework by eliminating any input-dependent preprocessing: a single shared encoder directly processes one channel at a time, thereby satisfying the strict channel-free definition introduced in Section~\ref{sec:problem_setting}.

    \item[(ii)] \textbf{Channel semantics.} 
    \cite{Hasegawa2024} organized channels using grouping heuristics, such as per-sensor, per-axis, per-type, or data-driven clustering via Jensen--Shannon divergence. In contrast, the present work explicitly injects semantic information about each channel through metadata conditioning based on body location, left/right, sensor type, and axis. This provides a principled way to recover, within a late-fusion architecture, part of the structural information that early-fusion models can implicitly exploit through fixed channel order.

    \item[(iii)] \textbf{Training objective.} 
    \cite{Hasegawa2024} employed a single cross-entropy loss on the aggregated representation. The present work extends this objective by introducing a combination loss that jointly supervises the fused prediction and the channel-wise auxiliary predictions. This encourages both global consistency and channel-wise discriminability, yielding an ensemble-style training effect within a shared channel-free model.

    \item[(iv)] \textbf{Evaluation scope.} 
    \cite{Hasegawa2024} focused mainly on clean accuracy on PAMAP2 together with a simulated single-IMU-omission setting. In the present work, we considerably broaden the evaluation scope by including systematic channel-perturbation robustness (shuffle, missing, joint, and continuous intensity sweeps), cross-dataset evaluation on six benchmarks, backbone generalization across five architectures, and cross-dataset transfer learning experiments.
\end{enumerate}

Building on these prior efforts, this study focuses on \textbf{strict} channel-free HAR, in which a single shared model performs inference even when the number of input sensors or channels differs across samples, without input-layer redesign. To this end, we organize the design space of integration mechanisms that satisfy the channel-free constraint from the perspectives of fusion timing (EF/MF/LF) and fusion operation, and connect this space to permutation-invariant set-processing architectures such as DeepSets~\cite{NIPS2017_f22e4747}. On this basis, we adopt LF, which is naturally compatible with permutation-invariant processing of channel sets, while explicitly injecting structural information through metadata. In this sense, the main contributions of this study are: (a) formulating strict channel-free HAR as a new problem setting, (b) explicitly designing the architecture in consideration of fusion-related inductive biases, and (c) demonstrating that such a design improves not only flexibility but also recognition performance and robustness.

\section{Channel-Free HAR Model}
\subsection{Problem Setting}
\label{sec:problem_setting}
In this study, we define strict channel-free HAR as a recognition setting that satisfies the following conditions:
\begin{itemize}
    \item The input is represented as $\mathcal{X}\in\mathbb{R}^{b\times c\times l}$, where $b$, $c$, and $l$ denote the batch size, the number of input channels, and the window length, respectively. Here, both $c$ and $l$ are allowed to vary.
    \item The prediction is permutation-invariant with respect to the input channels.
    \item A single shared model can process unseen sensors and unseen channel configurations without introducing sensor-specific input layers or dataset-specific channel templates.
    \item General preprocessing required for sensor time series, such as sampling-rate unification and normalization, is allowed.
\end{itemize}

Under this definition, the input is not treated as a fixed-length ordered vector, but rather as a variable-length unordered set of channels, $\{\mathcal{X}_i\}_{i=1}^{c}$. This formulation is suitable for heterogeneous sensor configurations, where the number, order, and semantics of channels may differ across datasets and deployment settings. It also provides a basis for cross-dataset transfer and, in principle, for data-mixtured pretraining across multiple HAR datasets. 

\subsection{Inductive Biases of Fusion Strategies Under the Channel-Free Constraint}

In a channel-free model, the input $\mathcal{X}\in\mathbb{R}^{b\times c\times l}$ must be processed in a permutation-invariant manner. In the language of permutation-invariant set modeling~\cite{NIPS2017_f22e4747}, any channel-free classifier $F$ must factorize as $F(\{\mathcal{X}_i\}_{i=1}^{c}) = \rho\bigl(\bigoplus_{i=1}^{c}\phi(\mathcal{X}_i)\bigr)$ for some per-channel encoder $\phi$, commutative aggregator $\bigoplus$, and classifier $\rho$. This decomposition clarifies the design space: the shared encoder corresponds to $\phi$, mean pooling corresponds to a specific choice of $\bigoplus$, and the fused classifier corresponds to $\rho$. However, the decomposition itself does not dictate when the per-channel encoding should end and aggregation should begin. Different aggregation stages, such as the raw-signal, intermediate-feature, and final-feature levels, introduce different inductive biases. As illustrated in \figref{fig:fusion}, fusion strategies can be categorized as follows: (i) Early Fusion (EF), which aggregates the channels immediately after input and then feeds the fused signal into a single encoder; (ii) Middle Fusion (MF), which introduces channel interactions in intermediate layers; and (iii) Late Fusion (LF), which first encodes each channel independently and then integrates the resulting channel-wise features.

Their practical differences are summarized in \tabref{table:fusion_strength}. EF places a strong inductive bias on cross-channel interaction, but its encoder is tightly coupled to the fused input structure, making adaptation to changed channel configurations difficult. MF relaxes this coupling to some extent, but additional sensor-specific components may still be required depending on the design. By contrast, LF with a shared encoder is most naturally suited to channel-free HAR: newly added channels can be processed by the same backbone without modification, and permutation-invariant aggregation can then be applied afterward.

These observations motivate the proposed design. We adopt an LF-style architecture as the basis for channel-free processing, explicitly reintroduce channel semantics through metadata such as body location, sensor type, and axis, and optimize both channel-wise and fused predictions so that individual channels remain discriminative while the final prediction is jointly improved.

\begin{table}[t]
    \caption{Functional strengths of EF, LF, and the proposed model}
    \label{table:fusion_strength}
    \centering
    \begin{tabular}{lccc} \hline\hline
    Functional property & EF & LF & Ours \\ \hline
    Strong cross-channel interaction bias & \checkmark &  &  \\
    Ability to exploit channel semantics & \checkmark &  & \checkmark \\
    Permutation-invariant aggregation &  & \checkmark & \checkmark \\
    Ensemble-style prediction &  & \checkmark & \checkmark \\ \hline\hline
    \end{tabular}
\end{table}

\subsection{Proposed Method}

\begin{figure}[t]
    \centering
    \includegraphics[width=1.0\linewidth]{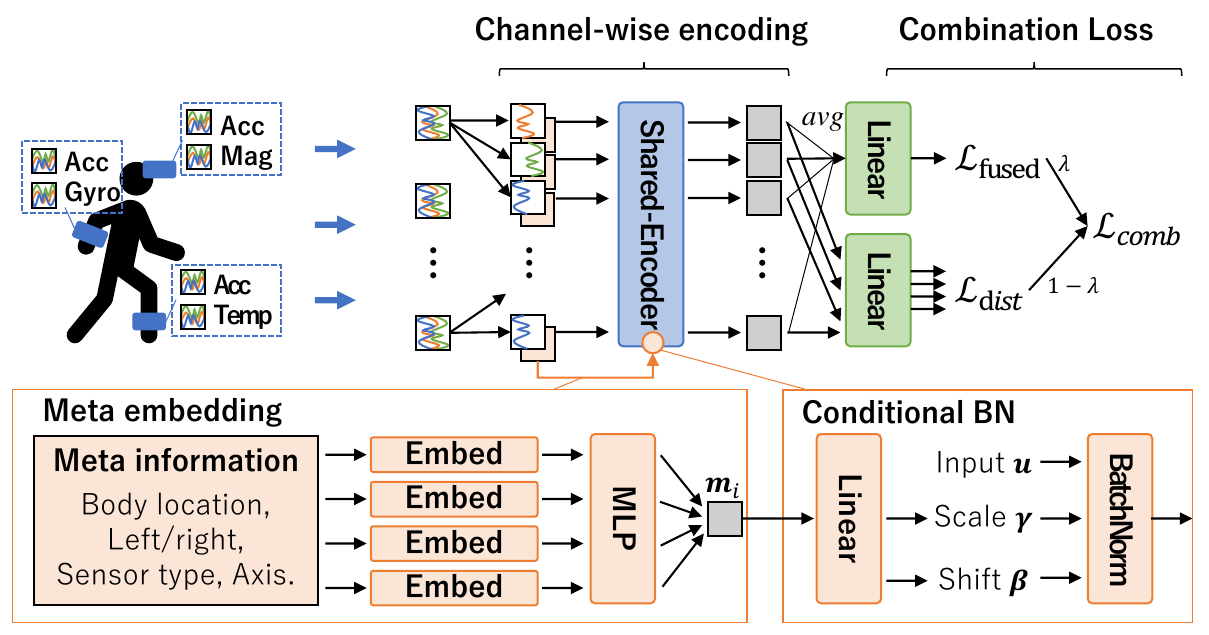}
    \caption{Architecture of the proposed channel-free HAR model. Each channel is independently processed by a shared encoder, and channel metadata are embedded and injected through conditional batch normalization. The resulting channel-wise features are aggregated by mean pooling for fused prediction, while auxiliary channel-wise predictions are also imposed. The model is trained using the combination loss $\mathcal{L}_{\mathrm{comb}}$, which combines the fused loss $\mathcal{L}_{\mathrm{fused}}$ and the channel-wise loss $\mathcal{L}_{\mathrm{dist}}$.}
    \label{fig:ours}
\end{figure}

As illustrated in \figref{fig:ours}, the proposed method consists of three components: (1) channel-wise encoding with a shared encoder, (2) metadata conditioning through meta embedding and conditional batch normalization, and (3) joint supervision of fused and channel-wise predictions through a combination loss. The first and third components satisfy the channel-free constraint, while the second explicitly injects structural information that would otherwise be lost in a purely channel-independent LF design.

\subsubsection{Channel-Wise Encoding}

Let the input be denoted by $\mathcal{X}\in\mathbb{R}^{b\times c\times l}$, and let the time series of the $i$-th channel be denoted by $\mathcal{X}_i\in\mathbb{R}^{b\times 1\times l}$. In the proposed method, each channel is processed independently by a shared encoder $f_{\theta}(\cdot)$, yielding a channel-wise feature vector
\begin{equation}
\mathbf{z}_i = f_{\theta}(\mathcal{X}_i).
\end{equation}
Because the same encoder is reused for all channels, the backbone itself does not need to be redesigned when the number or composition of input channels changes.

In implementation, this channel-wise encoding can be realized efficiently by reshaping the input into
\begin{equation}
\tilde{\mathcal{X}}\in\mathbb{R}^{(b\cdot c)\times 1\times l}
\end{equation}
and feeding it to the shared encoder in parallel. In this way, the model treats the input as a set of channels rather than as a fixed sensor template, which is consistent with the strict channel-free setting. Because the same encoder is reused for all channels, the backbone itself does not need to be redesigned when the number or composition of input channels changes. Although the computational cost still depends on the number of input channels, the shared-encoder design enables efficient parallel implementation. Its practical cost is examined in the experimental section.

\subsubsection{Meta Embedding and Conditional Batch Normalization}

A purely channel-wise LF model is flexible with respect to channel configuration, but it may lose structural information that EF implicitly exploits under fixed channel arrangements. To compensate for this loss, the proposed method explicitly injects channel semantics through metadata conditioning.

For each channel $i$, metadata are represented by four discrete identifiers: body location $p_i$, left/right information $\ell_i$, sensor type $s_i$, and sensor axis $a_i$. Their embeddings are denoted by $\mathbf{e}^{(p)}_i$, $\mathbf{e}^{(\ell)}_i$, $\mathbf{e}^{(s)}_i$, and $\mathbf{e}^{(a)}_i$, respectively. These embeddings are concatenated and transformed by an MLP to produce a channel-specific metadata vector $\mathbf{m}_i$:
\begin{equation}
\mathbf{m}_i :=
\mathrm{LN}\!\Bigl(
f_m\!\bigl(
\mathrm{cat}(
\mathbf{e}^{(p)}_i,\,
\mathbf{e}^{(\ell)}_i,\,
\mathbf{e}^{(s)}_i,\,
\mathbf{e}^{(a)}_i
)
\bigr)
\Bigr).
\end{equation}
Here, $\mathrm{cat}(\cdot)$ denotes vector concatenation, $f_m(\cdot)$ denotes an MLP, and $\mathrm{LN}(\cdot)$ denotes layer normalization. Each embedding uses ID$=0$ as a padding token. If metadata are unavailable for a channel, the corresponding embedding is treated as padding so that the model can still operate when metadata are partially missing.

The resulting metadata vector $\mathbf{m}_i$ is injected into the shared encoder through Conditional Batch Normalization (CBN) layers~\cite{Harm2017CBN}. Let $\mathbf{u}$ denote an intermediate activation of channel $i$ in the backbone. Then, CBN first normalizes $\mathbf{u}$ by BatchNorm and subsequently modulates the normalized activation using a scale and bias derived from $\mathbf{m}_i$:
\begin{equation}
\mathrm{CBN}(\mathbf{u};\mathbf{m}_i)
=
\bigl(1+\lambda_{\gamma}\tanh(\gamma(\mathbf{m}_i))\bigr)\odot \mathrm{BN}(\mathbf{u})
+\beta(\mathbf{m}_i),
\label{eq:cbn}
\end{equation}
where $\gamma(\mathbf{m}_i)$ and $\beta(\mathbf{m}_i)$ are obtained from a linear projection of $\mathbf{m}_i$, and $\lambda_{\gamma}$ is a coefficient that limits the modulation strength. Compared with the standard CBN parameterization $\gamma(\mathbf{m})\odot\mathrm{BN}(\mathbf{u})+\beta(\mathbf{m})$, Eq.~\eqref{eq:cbn} parameterizes the scale as a bounded perturbation from the identity map: when the metadata-derived signal is weak or unreliable, $\tanh(\gamma(\mathbf{m}_i))\approx 0$ yields a scale of $1$ and the layer degenerates to plain BatchNorm. This design choice keeps the backbone stable when metadata are noisy, corrupted, or entirely absent (e.g., when ID$=0$ padding is used), and is consistent with the graceful degradation behavior observed in the meta-inconsistent perturbation experiments of Section~\ref{sec:perturbation}. In other words, the metadata vector does not replace the signal representation itself, but instead gently conditions the intermediate feature normalization process. By embedding CBN into the shared backbone, the proposed method explicitly provides the channel semantics that EF would otherwise exploit only implicitly under a fixed channel order.

\subsubsection{Fused Prediction and Channel-Wise Auxiliary Prediction}

After channel-wise encoding, the shared encoder produces a feature vector $\mathbf{z}_i\in\mathbb{R}^{D}$ for each channel. These channel-wise features are used in two ways.

First, for the fused prediction, the features are aggregated by simple mean pooling:
\begin{equation}
\bar{\mathbf{z}}=\frac{1}{C}\sum_{i=1}^{C}\mathbf{z}_i,
\end{equation}
and the fused classifier $h_{\theta}$ produces the final prediction
\begin{equation}
\hat{y}_{\mathrm{fused}} = h_{\theta}(\bar{\mathbf{z}}).
\end{equation}
This aggregation is permutation-invariant and does not rely on a learnable fusion operator, which makes it naturally consistent with the channel-free constraint.

Second, in order to preserve the discriminability of individual channel features, an auxiliary classifier $h_{\psi}$ is applied to each channel feature:
\begin{equation}
\hat{y}_i = h_{\psi}(\mathbf{z}_i), \qquad i=1,\dots,C.
\end{equation}
Thus, the proposed method uses both a fused prediction based on the mean-aggregated feature and channel-wise auxiliary predictions based on each $\mathbf{z}_i$.

\subsubsection{Combination Loss}

The fused prediction is optimized by the fused loss
\begin{equation}
\mathcal{L}_{\mathrm{fused}}
=
\mathcal{L}_{\mathrm{ce}}(y,\hat{y}_{\mathrm{fused}}),
\end{equation}
where $y$ denotes the ground-truth label. If only this loss is used, however, the encoder is optimized solely for the final fused prediction, and the discriminability of each channel-wise feature is not explicitly maintained.

To address this issue, the channel-wise auxiliary predictions are supervised by
\begin{equation}
\mathcal{L}_{\mathrm{dist}}
=
\frac{1}{C}\sum_{i=1}^{C}\mathcal{L}_{\mathrm{ce}}(y,\hat{y}_i).
\end{equation}
This loss encourages each channel feature to retain predictive information individually, while the fused loss directly optimizes the final prediction after aggregation.

Finally, the proposed method is trained using the combination loss
\begin{equation}
\mathcal{L}_{\mathrm{comb}}
=
\lambda \mathcal{L}_{\mathrm{fused}} + (1-\lambda)\mathcal{L}_{\mathrm{dist}},
\qquad \lambda\in[0,1].
\label{eq:comb}
\end{equation}
Here, $\lambda=1$ corresponds to using only the fused loss, whereas $\lambda=0$ corresponds to using only the channel-wise auxiliary loss. By combining the two objectives, the proposed method simultaneously exploits the robustness of permutation-invariant mean aggregation and the discriminability of channel-wise representations. In the experiments, $\lambda$ is set to 0.5 by default, so that the fused and channel-wise losses are equally weighted, and its sensitivity is analyzed separately in Section~\ref{sec:sensitivity}.

\section{Experiments}
\label{sec:experiments}

\subsection{Experimental Setup}
\label{sec:setup}

\subsubsection{Datasets}
We evaluate on six publicly available HAR datasets (PAMAP2~\cite{Reiss2012a}, DSADS~\cite{Altun2010-kp}, MHEALTH~\cite{Banos2014-uc}, UCI-HAR~\cite{Anguita2013-qb}, UniMiB SHAR~\cite{Daniela2017}, and WISDM~\cite{Kwapisz2011}) covering a wide range of sensor configurations and activity types (Table~\ref{table:datasets}).All signals are resampled to \SI{100}{Hz}, segmented into 256-sample windows (2.56 s), and standardization based on the training set; no other preprocessing is applied. While PAMAP2 originally contains 52 channels, we excluded the orientation measurement data due to reported issues, resulting in our analysis using 40 remaining channels\footnote{PAMAP2 Physical Activity Monitoring: \url{http://archive.ics.uci.edu/dataset/231/pamap2+physical+activity+monitoring}}.

\subsubsection{Evaluation Protocol and Metrics}
Leave-One-Subject-Out cross-validation (LOSO-CV) is used throughout to obtain subject-independent performance estimates. Experiments on the PAMAP2 dataset are repeated with five independent random seeds, and the results are reported as the mean $\pm$ standard deviation over all folds (subjects) and seeds. The primary metrics are macro-averaged accuracy and macro-averaged F1-score.

Beyond clean accuracy, we report three robustness metrics:
\begin{itemize}
    \item \textbf{Clean}: Test score under the original input configuration without perturbation.
    \item \textbf{Shuffle (Shfl)}: Test inputs with the channel order randomly permuted, while meta-information is kept aligned with the permuted channels.
    \item \textbf{Missing (Miss)}: A randomly selected fraction (50\,\%) of channels is zeroed out, and the corresponding meta tokens are masked.
    \item \textbf{Shfl+Miss}: Both perturbations are applied simultaneously.
\end{itemize}

\subsubsection{Models and Training Details}
All models share the same 1D ResNet10 backbone~\cite{He2016resnet}, consisting of four residual blocks, a base channel size of 32, and a convolution kernel size of 3.  We compare the following variants. Note that the EF and MF variants are implemented here with slot-attention~\cite{Locatello2020SlotoAttn} so that they can accommodate a variable number of input channels; this was necessary because most existing EF/MF HAR architectures assume a fixed channel count and therefore cannot serve as direct baselines under the strict channel-free setting. Accordingly, the EF/MF results in this paper should be read as representative channel-free adaptations of early and middle fusion, rather than as reproductions of any specific prior method.

\begin{itemize}
  \item \textbf{Baseline}: Common \textbf{channel-fixed} model trained with cross-entropy only.
  \item \textbf{EF}: Early-fusion with slot attention~\cite{Locatello2020SlotoAttn}. $K$ learnable slot queries cross-attend to per-channel waveform summaries (mean, std, max, min, energy), yielding soft assignment weights $\mathbf{A}\in\mathbb{R}^{K\times C}$. Raw waveforms are mixed as $\mathbf{y}_k = \sum_{c=1}^{C} A_{kc}\,\mathbf{x}_c$, reducing $C$ variable channels to $K{=}16$ fixed virtual channels. The $K$ channels are then processed independently by the per-channel backbone, followed by LF mean pooling and a linear classifier.
  \item \textbf{MF}: Middle-fusion with slot attention. The same slot-attention assignment is applied to intermediate feature maps after the first residual stage, mixing $\mathbf{h}_c\in\mathbb{R}^{F\times L}$ into $K$ virtual channels. The $K$ channels are processed independently through the remaining backbone stages, then aggregated by LF mean pooling.
  \item \textbf{LF}: Late-fusion with per-channel backbone features aggregated by masked mean pooling over valid channels; the pooled representation is then classified by a single linear layer trained with cross-entropy loss.
  \item \textbf{Ours}: Late-fusion with meta-embedding trained by the combination loss Eq.~\eqref{eq:comb}.
\end{itemize}

\subsubsection{Training Details}
All models are trained with Adam (lr $= 10^{-3}$, weight decay $= 0$), cosine annealing (T\textsubscript{max} $=$ epochs), batch size 64, and a fixed 50 epochs per fold without early stopping. Standardization (zero-mean, unit-variance per channel, computed on the training split of each fold) is applied to all inputs.

\subsection{Main Comparison on PAMAP2 Under Channel Perturbations}
\label{sec:pamap_main}
\begin{figure}[bp]
  \centering
  \includegraphics[width=0.8\linewidth]{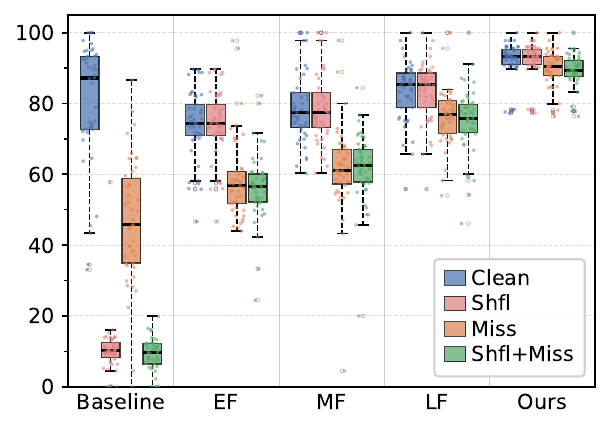}
  \caption{Accuracy (\%) distributions on PAMAP2 for Baseline, EF, MF, LF, and the proposed method under clean and perturbed conditions. Results are summarized as boxplots over LOSO-CV folds and five random trials.}
  \label{fig:pamap_main}
\end{figure}

Fig.~\ref{fig:pamap_main} presents the main comparison on PAMAP2. Baseline achieved strong performance under the clean condition, but its accuracy dropped substantially under perturbed conditions. This indicates that the conventional fixed-channel design works well only when the channel configuration is strictly preserved. In contrast, all channel-free methods generally suppressed the adverse effects of perturbations, demonstrating the advantage of removing the fixed-channel assumption in heterogeneous sensor environments. Notably, LF already achieved clean-condition performance competitive with the Baseline while being much more robust to perturbations, suggesting that late fusion itself is a strong channel-free baseline for HAR. Building on this, the proposed method further improved both performance and stability: by incorporating meta-embedding and the combination loss, it not only maintained strong robustness under perturbed conditions but also outperformed the Baseline even under the clean condition, while showing reduced variance across folds and trials.

A clear difference was observed between Shfl and Miss. The Baseline was particularly vulnerable to Shfl, because channel permutation breaks the fixed correspondence between channel positions and sensor semantics assumed during training. By contrast, the channel-free methods were largely insensitive to Shfl, as their order-agnostic design reduces dependence on channel order. For Miss, some performance degradation remained even for channel-free methods, since missing channels directly reduce the available sensor information. Nevertheless, the degradation was still much smaller than that of the Baseline, indicating that channel-free models can flexibly exploit the remaining valid channels. Overall, these results show that channel-free modeling is highly effective against sensitivity to channel order and also substantially mitigates the impact of information loss. 

\subsection{Ablation Study}
\label{sec:ablation}
\begin{table}[b]
  \centering
  \caption{Ablation study on PAMAP2 using LOSO-CV with 5 trials. Results are reported as mean\,\textpm\,std (\%) over 5 trial-level means, where each trial-level mean is averaged over 9 subjects. Best result per column in \textbf{bold}.}
  \label{tab:ablation}
  \begin{tabular}{l cccc}
    \toprule
    \textbf{Model} & \textbf{Acc} & \textbf{F1} & \textbf{Miss-Acc} & \textbf{Shfl+Miss} \\
    \midrule
    Baseline & 80.3\,{\tiny $\pm$\,1.1} & 61.6\,{\tiny $\pm$\,0.9} & 46.6\,{\tiny $\pm$\,4.5} & 8.9\,{\tiny $\pm$\,1.2} \\
    LF & 83.4\,{\tiny $\pm$\,2.1} & 66.8\,{\tiny $\pm$\,1.1} & 76.6\,{\tiny $\pm$\,1.8} & 75.2\,{\tiny $\pm$\,3.3} \\
    LF + $\mathcal{L}_{\mathrm{dist}}$ & 86.6\,{\tiny $\pm$\,0.3} & 71.1\,{\tiny $\pm$\,0.3} & 85.3\,{\tiny $\pm$\,0.8} & 85.2\,{\tiny $\pm$\,1.0} \\
    LF + $\mathcal{L}_{\mathrm{comb}}$ & 89.8\,{\tiny $\pm$\,0.7} & 72.5\,{\tiny $\pm$\,0.5} & 86.0\,{\tiny $\pm$\,1.1} & 85.3\,{\tiny $\pm$\,1.1} \\
    LF + $\mathcal{L}_{\mathrm{comb}}$ + Meta & \textbf{92.0}\,{\tiny $\pm$\,0.2} & \textbf{74.5}\,{\tiny $\pm$\,0.3} & \textbf{89.6}\,{\tiny $\pm$\,1.0} & \textbf{89.1}\,{\tiny $\pm$\,0.4} \\
    \bottomrule
  \end{tabular}
\end{table}
Table~\ref{tab:ablation} presents an incremental ablation of the proposed LF-based design. First, replacing the channel-fixed Baseline with the plain LF model already improved both clean performance and perturbation robustness, indicating that the channel-free LF structure itself is the main source of robustness. Second, adding auxiliary supervision further improved all metrics. Although both $\mathcal{L}_{\mathrm{dist}}$ and $\mathcal{L}_{\mathrm{comb}}$ were effective, $\mathcal{L}_{\mathrm{comb}}$ consistently achieved better clean performance while preserving strong robustness, supporting the use of balanced supervision between aggregated and channel-wise predictions. Finally, adding meta-conditioning provided further gains across all metrics and also reduced variance across trials. These results show that the final performance is obtained through a clear stepwise improvement from LF to LF + $\mathcal{L}_{\mathrm{comb}}$ and then to LF + $\mathcal{L}_{\mathrm{comb}}$ + Meta.

\subsection{Efficiency Analysis}
\label{sec:efficiency}
To quantify the computational overhead of the proposed design, we compared model efficiency on PAMAP2-format inputs ($C=40$, $L=256$, batch size = 4) using dummy data. Table~\ref{tab:latency} reports the number of parameters, MACs per sample, inference time, and training time. All measurements were obtained in a WSL2-based Ubuntu environment with an Intel Core i9-13900KF CPU, 19\,GiB visible system memory, an NVIDIA GeForce RTX 4090 GPU (24\,564\,MiB), Python 3.12, and PyTorch 2.6.0 with CUDA 12.4.

\begin{table}[t]
  \centering
  \caption{Efficiency comparison on PAMAP2-format inputs ($C=40$, $L=256$, batch size = 4). MACs are reported per sample, while inference and training times are wall-clock averages measured on GPU.}
  \label{tab:latency}
  \begin{tabular}{l r r r r}
    \toprule
    \textbf{Model} & \textbf{\#Params} & \textbf{MACs} & \textbf{Infer} & \textbf{Train}  \\
    & (M) & (M/sample) & (ms/batch) & (ms/step) \\
    \midrule
    Baseline & 0.45 & 6.2 & 0.72 & 3.16 \\
    EF & 0.46 & 81.9 & 1.51 & 5.15 \\
    MF & 0.47 & 204.9 & 1.52 & 5.22 \\
    LF & 0.44 & 204.7 & 0.86 & 3.56 \\
    LF+$\mathcal{L}_{\mathrm{comb}}$ & 0.45 & 204.7 & 0.89 & 4.09 \\
    LF+$\mathcal{L}_{\mathrm{comb}}$+Meta & 0.59 & 208.7 & 2.14 & 7.38 \\
    \bottomrule
  \end{tabular}
\end{table}
\begin{figure}[b]
  \centering
  \includegraphics[width=1.\linewidth]{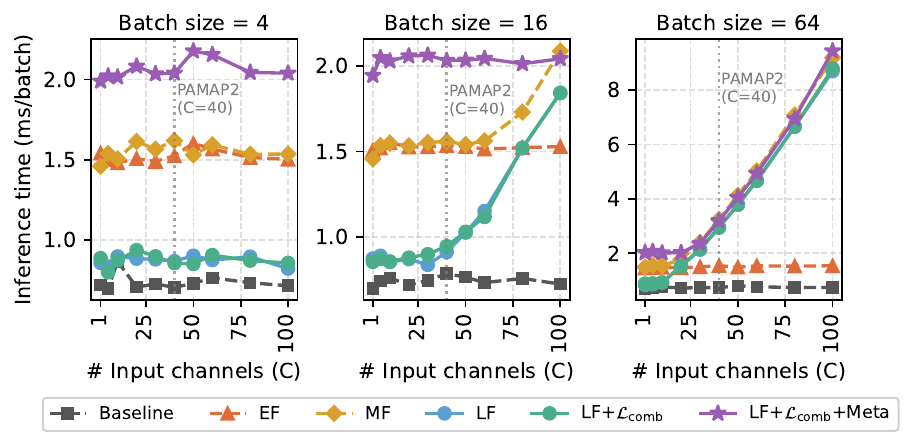}
  \caption{Inference time as a function of the number of input channels for different batch sizes. LF-based models remain efficient at moderate channel counts, while the cost increase becomes more visible as the number of channels and batch size grow.}
  \label{fig:channel_scaling}
\end{figure}

As shown in Table~\ref{tab:latency}, the proposed LF+$\mathcal{L}_{\mathrm{comb}}$+Meta model is more expensive than the channel-fixed Baseline, mainly due to the channel-wise encoding and meta-conditioning module. However, the increase in parameter count remains moderate (0.45M to 0.59M), and the model still runs in the millisecond range for both inference and training. LF and LF+$\mathcal{L}_{\mathrm{comb}}$ also maintain relatively low latency despite their larger MAC counts, suggesting that shared per-channel processing can be executed efficiently on GPU. In practice, the additional overhead of the proposed model appears acceptable given the substantial gains in clean accuracy, robustness, and transferability shown in the preceding and following sections. The measurements reported here target a workstation-class GPU; a systematic evaluation on resource-constrained edge devices (e.g., ARM CPUs or mobile NPUs), which is more directly relevant to on-device IoT deployment, is left as future work.

Fig.~\ref{fig:channel_scaling} illustrates how inference latency varies with the number of input channels for different batch sizes. For LF-based models, the latency remains nearly flat at small batch sizes, suggesting that the additional per-channel computation is largely absorbed by GPU parallelism. In contrast, when the batch size and channel count become large, the latency increases much more sharply, indicating that memory pressure becomes the dominant factor.

\subsection{Robustness Analysis}
\label{sec:robustness}
\begin{table*}[t]
  \centering
  \caption{Cross-dataset results (LOSO-CV). \textbf{Clean}: standard test accuracy; \textbf{Pert}: accuracy under simultaneous channel shuffle and 50\,\% drop. PAMAP2: mean$\,\pm\,$std over 5 independent trials; others: mean$\,\pm\,$std over subjects (single trial). Best result per column in \textbf{bold}.}
  \label{tab:dataset_robustness}
  \begin{tabular}{l r r r r r r r r r r r r}
    \toprule
    & \multicolumn{2}{c}{PAMAP2 (40ch)} & \multicolumn{2}{c}{mHealth (23ch)} & \multicolumn{2}{c}{DSADS (45ch)} & \multicolumn{2}{c}{UCI-HAR (9ch)} & \multicolumn{2}{c}{WISDM (3ch)} & \multicolumn{2}{c}{UniMiB (3ch)} \\
    \cmidrule(lr){2-3} \cmidrule(lr){4-5} \cmidrule(lr){6-7} \cmidrule(lr){8-9} \cmidrule(lr){10-11} \cmidrule(lr){12-13}
    \textbf{Model} & {\small Clean} & {\small Pert} & {\small Clean} & {\small Pert} & {\small Clean} & {\small Pert} & {\small Clean} & {\small Pert} & {\small Clean} & {\small Pert} & {\small Clean} & {\small Pert} \\
    \midrule
    Baseline & 80.3{\tiny$\pm$1.1} & 8.9{\tiny$\pm$1.2} & 91.9{\tiny$\pm$5.5} & 15.7{\tiny$\pm$4.6} & 84.9{\tiny$\pm$6.7} & 7.4{\tiny$\pm$4.6} & 95.4{\tiny$\pm$6.2} & 33.5{\tiny$\pm$6.4} & 86.3{\tiny$\pm$18.3} & 37.5{\tiny$\pm$11.3} & 77.6{\tiny$\pm$10.6} & 49.2{\tiny$\pm$12.6} \\
    EF & 73.9{\tiny$\pm$1.5} & 56.4{\tiny$\pm$3.4} & 78.4{\tiny$\pm$9.9} & 67.1{\tiny$\pm$7.6} & 47.0{\tiny$\pm$11.2} & 38.0{\tiny$\pm$8.1} & 90.7{\tiny$\pm$5.5} & 75.1{\tiny$\pm$5.9} & 82.8{\tiny$\pm$18.1} & 72.1{\tiny$\pm$14.5} & 73.8{\tiny$\pm$8.3} & 68.4{\tiny$\pm$8.8} \\
    MF & 78.8{\tiny$\pm$1.1} & 61.7{\tiny$\pm$5.1} & 88.7{\tiny$\pm$5.0} & 73.3{\tiny$\pm$11.7} & 77.4{\tiny$\pm$6.8} & 49.7{\tiny$\pm$7.0} & 91.1{\tiny$\pm$7.4} & 73.2{\tiny$\pm$8.5} & 84.6{\tiny$\pm$20.3} & 66.1{\tiny$\pm$17.6} & 78.3{\tiny$\pm$8.6} & 61.9{\tiny$\pm$11.2} \\
    LF & 83.4{\tiny$\pm$2.1} & 75.2{\tiny$\pm$3.3} & 91.1{\tiny$\pm$7.3} & 86.7{\tiny$\pm$9.1} & 88.4{\tiny$\pm$3.9} & 81.3{\tiny$\pm$4.6} & 91.7{\tiny$\pm$5.9} & 82.2{\tiny$\pm$6.3} & 85.1{\tiny$\pm$15.7} & 75.7{\tiny$\pm$14.7} & 79.8{\tiny$\pm$7.9} & 74.0{\tiny$\pm$7.8} \\
    Ours & \textbf{92.0{\tiny$\pm$0.2}} & \textbf{89.1{\tiny$\pm$0.4}} & \textbf{98.3{\tiny$\pm$1.8}} & \textbf{93.8{\tiny$\pm$7.7}} & \textbf{92.8{\tiny$\pm$4.8}} & \textbf{89.3{\tiny$\pm$5.2}} & \textbf{95.5{\tiny$\pm$4.5}} & \textbf{87.1{\tiny$\pm$5.1}} & \textbf{88.0{\tiny$\pm$15.0}} & \textbf{82.8{\tiny$\pm$13.5}} & \textbf{80.8{\tiny$\pm$7.3}} & \textbf{76.0{\tiny$\pm$7.5}} \\
    \bottomrule
  \end{tabular}
\end{table*}

\begin{table*}[t]
  \centering
  \caption{Backbone architecture comparison on PAMAP2 LOSO-CV. Clean: standard test accuracy; Pert: accuracy under simultaneous channel shuffle and 50\,\% drop. ResNet: mean$\,\pm\,$std over 5 trials; others: mean$\,\pm\,$std over 9 subjects (1 trial). Best per column in \textbf{bold}.}
  \label{tab:backbone_comparison}
  \begin{tabular}{l r r r r r r r r r r}
    \toprule
    & \multicolumn{2}{c}{ResNet10~\cite{He2016resnet}} & \multicolumn{2}{c}{TCN~\cite{Lea_2017_CVPR}} & \multicolumn{2}{c}{VGG16~\cite{Simonyan2014}} & \multicolumn{2}{c}{MobileNetV2~\cite{Sandler_2018_CVPR}} & \multicolumn{2}{c}{EfficientNetB0~\cite{pmlr-v97-tan19a}} \\
    \cmidrule(lr){2-3} \cmidrule(lr){4-5} \cmidrule(lr){6-7} \cmidrule(lr){8-9} \cmidrule(lr){10-11}
    \textbf{Method} & {\small Clean} & {\small Pert} & {\small Clean} & {\small Pert} & {\small Clean} & {\small Pert} & {\small Clean} & {\small Pert} & {\small Clean} & {\small Pert} \\
    \midrule
    Baseline & 80.3{\tiny$\pm$1.1} & 8.9{\tiny$\pm$1.2} & 78.1{\tiny$\pm$20.6} & 10.0{\tiny$\pm$5.4} & 79.2{\tiny$\pm$16.4} & 11.4{\tiny$\pm$5.8} & 82.1{\tiny$\pm$18.4} & 14.4{\tiny$\pm$6.2} & 81.7{\tiny$\pm$18.1} & 12.3{\tiny$\pm$6.9} \\
    LF + $\mathcal{L}_{\mathrm{comb}}$ & 89.8{\tiny$\pm$0.7} & 85.3{\tiny$\pm$1.1} & 86.3{\tiny$\pm$5.5} & 79.2{\tiny$\pm$4.9} & 90.6{\tiny$\pm$3.7} & 86.7{\tiny$\pm$3.0} & 90.5{\tiny$\pm$4.5} & 85.0{\tiny$\pm$2.6} & 88.2{\tiny$\pm$3.6} & 85.6{\tiny$\pm$4.3} \\
    LF + $\mathcal{L}_{\mathrm{comb}}$ + Meta & \textbf{92.0{\tiny$\pm$0.2}} & \textbf{89.1{\tiny$\pm$0.4}} & \textbf{89.2{\tiny$\pm$6.3}} & \textbf{86.5{\tiny$\pm$6.2}} & \textbf{91.5{\tiny$\pm$5.0}} & \textbf{89.4{\tiny$\pm$4.7}} & \textbf{91.1{\tiny$\pm$6.5}} & \textbf{88.1{\tiny$\pm$6.3}} & \textbf{91.6{\tiny$\pm$5.3}} & \textbf{89.2{\tiny$\pm$5.7}} \\
    \bottomrule
  \end{tabular}
\end{table*}

Tables~\ref{tab:dataset_robustness} and~\ref{tab:backbone_comparison} evaluate the robustness of the proposed method across both datasets and backbone architectures. In both analyses, a highly consistent tendency was observed. The channel-fixed Baseline was often competitive under the clean condition, but its performance degraded drastically under perturbation, indicating strong dependence on a fixed correspondence between channel positions and sensor semantics. By contrast, the channel-free methods were substantially more robust across all datasets and all backbones. It is worth noting that UniMiB SHAR exhibits a relatively small Baseline--Pert gap compared with other datasets because it contains only three accelerometer channels from a single location. As a result, channel shuffling only permutes the three axes of a single accelerometer, rather than disrupting relationships across multiple sensors or locations. Similarly, masking 50\,\% of the channels still preserves part of the same tri-axial signal, making the perturbation less severe than in datasets with richer sensor configurations. In other words, the apparent Baseline robustness on UniMiB is a consequence of low sensor heterogeneity, not of architectural robustness, which is itself consistent with our thesis that channel-free design matters most under heterogeneous sensor configurations.

Among the channel-free baselines, LF again showed the strongest balance between clean and perturbed performance, confirming that late fusion is a simple yet strong formulation for channel-free HAR. On top of this baseline, the proposed method achieved the best results on all datasets and for all backbone architectures under both clean and perturbed conditions. The gains were especially pronounced on datasets with richer sensor heterogeneity, while remaining consistent across substantially different encoders such as TCN~\cite{Lea_2017_CVPR}, VGG16~\cite{Simonyan2014}, MobileNetV2~\cite{Sandler_2018_CVPR}, and EfficientNetB0~\cite{pmlr-v97-tan19a}.

Overall, these results indicate that the superiority of the proposed method is not specific to PAMAP2, nor to the ResNet-based backbone used in the main experiments. Instead, the proposed channel-free design generalizes across diverse sensing conditions and model architectures, supporting its suitability for realistic heterogeneous HAR environments.

\subsection{Perturbation-Intensity Analysis}
\label{sec:perturbation}
\begin{figure*}[htbp]
  \centering
  \includegraphics[width=1.0\linewidth]{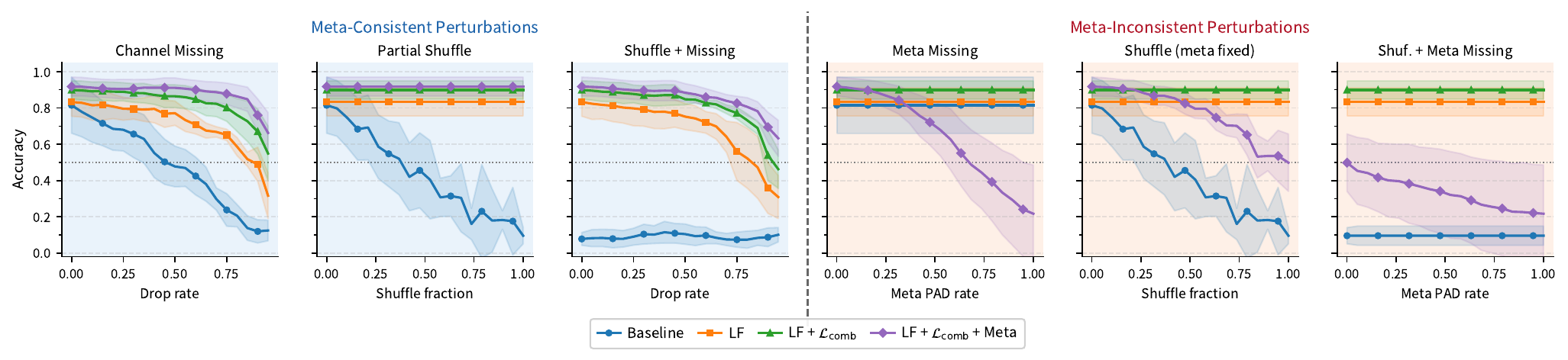}
  \caption{Accuracy as a function of perturbation intensity under six conditions on PAMAP2 (LOSO-CV; shaded: $\pm 1$ std). The left three panels apply meta-consistent perturbations, where signal and metadata are transformed jointly, whereas the right three panels apply meta-inconsistent perturbations, where signal--metadata alignment is intentionally broken. LF-based models remain robust under all meta-consistent conditions, while the meta-inconsistent conditions reveal that LF + $\mathcal{L}_{\mathrm{comb}}$ + Meta depends on metadata correctness: its accuracy drops relative to LF + $\mathcal{L}_{\mathrm{comb}}$ when metadata are corrupted or misaligned, whereas Baseline and LF are unaffected.}
  \label{fig:perturbation_sweep}
\end{figure*}

To further analyze the robustness of the proposed method, we conducted a perturbation sweep on PAMAP2 under LOSO-CV by continuously varying the perturbation intensity. We evaluated four models, namely Baseline, LF, LF + $\mathcal{L}_{\mathrm{comb}}$, and LF + $\mathcal{L}_{\mathrm{comb}}$ + Meta, under six perturbation conditions. These conditions can be grouped into two categories: meta-consistent perturbations, where the signal and metadata are transformed jointly (Channel Missing, Partial Shuffle, and Shuffle+Missing), and meta-inconsistent perturbations, where only the signal is modified so that the signal--metadata correspondence is intentionally broken (Meta Missing, Shuffle with fixed meta, and their combination). For each condition, the perturbation intensity parameter was varied from 0 to 1, and the mean and standard deviation of accuracy over LOSO-CV folds were recorded.

Fig.~\ref{fig:perturbation_sweep} shows the resulting accuracy curves. Under the meta-consistent conditions (left three panels), the proposed method remained highly robust even under strong perturbations. In the Channel Missing condition, it maintained over 80\% accuracy even when approximately 75\% of the channels were removed, indicating that the model can still exploit the remaining valid channels in highly degraded settings. In the Partial Shuffle condition, the accuracy remained nearly unchanged even when the shuffle fraction reached 1.0. This behavior reflects the order-invariant nature of the LF-based architecture, in which channel aggregation by mean pooling does not depend on channel order. A similar tendency was observed in the Shuffle+Missing condition, showing that LF-based models are robust even to combined perturbations.

In contrast, under the meta-inconsistent conditions (right three panels), LF + $\mathcal{L}_{\mathrm{comb}}$ + Meta showed a gradual performance drop as the perturbation intensity increased. This suggests that part of its discriminative ability relies on correctly aligned metadata, and such benefit diminishes once the correspondence between signal and metadata is corrupted. The Baseline and LF curves in the same panels appear flat simply because these models do not consume metadata at all, so perturbing the metadata stream cannot, by construction, affect their outputs; the right three panels therefore isolate the cost of introducing metadata conditioning and should not be read as a favorable comparison for Baseline or LF. Importantly, the degradation of LF + $\mathcal{L}_{\mathrm{comb}}$ + Meta remained limited in the low-intensity regime: when the Meta PAD rate or shuffle fraction was around 0.1, it still performed almost comparably to LF + $\mathcal{L}_{\mathrm{comb}}$. Moreover, since such signal--metadata inconsistency is less likely than ordinary channel missing or channel permutation in practical HAR deployments, the right three panels should be interpreted as a stress test rather than a standard operating condition. Overall, these results suggest that the proposed method is robust under realistic perturbations when metadata are correctly aligned, while the meta-free LF + $\mathcal{L}_{\mathrm{comb}}$ provides a reasonable fallback when metadata reliability is substantially compromised.

\subsection{Sensitivity Analysis}
\label{sec:sensitivity}

\begin{figure}[htbp]
  \centering
  \includegraphics[width=1.\linewidth]{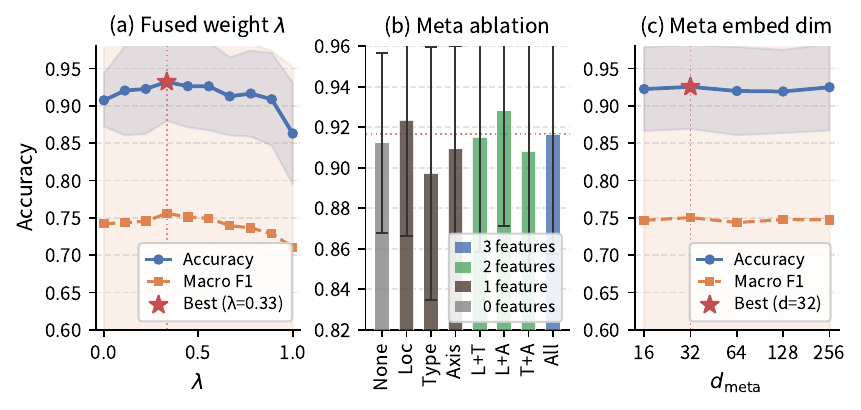}
  \caption{Sensitivity analysis on PAMAP2 (single trial, error bars: $\pm$1 std over subjects). (a) $\mathcal{L}_{\mathrm{comb}}$ mixing coefficient $\lambda$, (b) combinations of metadata features, and (c) meta-embedding dimension $d_{\mathrm{meta}}$.}
  \label{fig:sensitivity}
\end{figure}

We analyze the sensitivity of three hyperparameters in the proposed method on PAMAP2. \figref{fig:sensitivity}(a) shows the effect of the mixing coefficient $\lambda$ for $\mathcal{L}_{\mathrm{comb}}$. The best performance is obtained around $\lambda = 1/3$, but the accuracy remains nearly stable over a wide range, indicating that the method is not highly sensitive to the exact balance between the fused loss and the distributed loss. In contrast, the two extreme settings are clearly less effective. In particular, performance drops noticeably at $\lambda = 1$, confirming the importance of channel-wise auxiliary supervision. Since the optimal value of $\lambda$ may vary depending on the task~\cite{takama2026dynamic}, we adopt $\lambda = 0.5$ as the default setting in this paper to avoid over-optimizing for a specific condition.

\figref{fig:sensitivity}(b) examines the contribution of metadata components.
Among the combinations of location/side, sensor type, and axis information, the combination of location and axis yielded the highest performance.
Since this analysis is based on a single trial with PAMAP2 dataset, we avoid making a strong claim about the strict ordering among partial combinations. Nevertheless, the results suggest that sensor-type information may have a slightly negative effect. This may indicate a limitation of the CBN-based conditioning used in the proposed method, namely that it cannot uniformly absorb metadata of substantially different nature.

\figref{fig:sensitivity}(c) shows the effect of the meta-embedding dimension.
The performance peaks at $d_{\mathrm{meta}}=32$ and remains nearly unchanged for larger dimensions.
This indicates that the available metadata can be represented sufficiently well in a relatively small embedding space, and that larger embedding sizes provide only limited benefit.
In this paper, we adopt $d_{\mathrm{meta}}=64$ as the default setting.
This choice achieves performance comparable to the best case ($d_{\mathrm{meta}}=32$) while retaining sufficient representational capacity for generalization to a broader range of datasets.

\subsection{Cross-Dataset Transfer Learning}
\label{sec:transfer}
To evaluate whether the proposed channel-free design enables transfer learning across datasets with different channel configurations, we conducted cross-dataset pretraining experiments using PAMAP2 (40ch) as the target dataset. Since channel-free models do not require a fixed input channel structure, they can be pretrained on source datasets even when the number, order, and semantics of channels differ. We considered single-source pretraining on mHealth (23ch) and DSADS (45ch), as well as simple multitask pretraining on five datasets. After pretraining, the models were adapted to PAMAP2 using full fine-tuning (FT) or linear probing (LP). Throughout these experiments, the metadata vocabulary (body location, left/right, sensor type, and axis) is shared across datasets, so that, for instance, ``wrist accelerometer, x-axis'' is represented by the same set of IDs in the source and target datasets. This design is a practical consequence of the strict channel-free formulation: because the backbone does not depend on any dataset-specific input layer, the only bridge between source and target is the metadata embedding, which is required to be inter-operable. When a metadata category does not appear in the source dataset (e.g., PAMAP2 includes temperature but WISDM does not), the corresponding embedding is simply initialized and trained during target adaptation. 

\begin{table}[b]
  \centering
  \caption{Transfer learning results on PAMAP2 (40ch) using LOSO-CV. FT and LP denote full fine-tuning and linear probing, respectively. Results are reported as mean\,\textpm\,std (\%). Best result per column in \textbf{bold}.}
  \label{tab:transfer}
  \setlength{\tabcolsep}{4pt}
  \begin{tabular}{l l c c}
    \toprule
    \textbf{Source} & \textbf{Model} & \textbf{FT} & \textbf{LP} \\
    \midrule
    \multirow{3}{*}{Scratch}
      & LF & 85.4{\tiny$\pm$3.1} & -- \\
      & LF + $\mathcal{L}_{\mathrm{comb}}$ & 87.5{\tiny$\pm$7.0} & -- \\
      & LF + $\mathcal{L}_{\mathrm{comb}}$ + Meta & 91.3{\tiny$\pm$5.9} & -- \\
    \midrule
    \multirow{3}{*}{mHealth (23ch)}
      & LF & 85.5{\tiny$\pm$5.0} & 80.6{\tiny$\pm$5.7} \\
      & LF + $\mathcal{L}_{\mathrm{comb}}$ & 89.6{\tiny$\pm$5.3} & 83.5{\tiny$\pm$6.4} \\
      & LF + $\mathcal{L}_{\mathrm{comb}}$ + Meta & 91.9{\tiny$\pm$5.9} & 81.5{\tiny$\pm$3.5} \\
    \midrule
    \multirow{3}{*}{DSADS (45ch)}
      & LF & 86.2{\tiny$\pm$5.8} & 80.0{\tiny$\pm$6.9} \\
      & LF + $\mathcal{L}_{\mathrm{comb}}$ & 89.3{\tiny$\pm$5.1} & \textbf{84.1{\tiny$\pm$8.0}} \\
      & LF + $\mathcal{L}_{\mathrm{comb}}$ + Meta & \textbf{92.1{\tiny$\pm$6.0}} & 83.8{\tiny$\pm$5.4} \\
    \midrule
    \multirow{3}{*}{Multitask}
      & LF & 84.4{\tiny$\pm$5.3} & 82.0{\tiny$\pm$8.3} \\
      & LF + $\mathcal{L}_{\mathrm{comb}}$ & 87.7{\tiny$\pm$4.9} & \textbf{84.2{\tiny$\pm$6.7}} \\
      & LF + $\mathcal{L}_{\mathrm{comb}}$ + Meta & 90.7{\tiny$\pm$5.9} & 83.0{\tiny$\pm$5.8} \\
    \bottomrule
  \end{tabular}
\end{table}

Table~\ref{tab:transfer} shows that transfer from heterogeneous source datasets is feasible, which is itself an important advantage of channel-free modeling. In particular, pretraining on mHealth or DSADS generally improved the PAMAP2 performance over training from scratch, especially under full fine-tuning. The best result was obtained by LF + $\mathcal{L}_{\mathrm{comb}}$ + Meta pretrained on DSADS, which slightly outperformed the corresponding scratch setting.

At the same time, simple multitask pretraining over multiple datasets did not consistently improve performance beyond the best single-source transfer, and in fact the multitask FT result (90.7\,\%) is slightly below the scratch setting (91.3\,\%). In our setup, each source dataset used an independent classification head attached to the shared backbone, and the heads were discarded at target adaptation time. Under this configuration, two factors likely work against naive multitask pretraining: (i)~label spaces differ sharply across datasets (e.g., UniMiB includes fall-related classes not present in PAMAP2), meaning that gradients from the source heads are not fully aligned with the activities of interest; and (ii)~the mixing ratio of source datasets was not tuned, which may have allowed smaller or noisier datasets to disproportionately influence the shared representation. This suggests that, although channel-free modeling makes mixed-dataset pretraining structurally feasible, naively combining heterogeneous datasets is insufficient to fully exploit their potential. This interpretation is also consistent with recent work on HAR data-mixture optimization, which showed that cross-dataset pretraining across heterogeneous IMU datasets benefits from explicitly optimized dataset mixtures rather than naive combination~\cite{ban2025}. More principled approaches, such as label-space alignment, task-weighted sampling, or self-supervised objectives that are independent of label spaces, remain important directions for future work. Notably, the linear-probing results already place LF + $\mathcal{L}_{\mathrm{comb}}$ at a competitive level ($84.1$--$84.2,\%$ for DSADS and Multitask), indicating that the pretrained representation itself retains useful information even without fine-tuning. Overall, these results indicate that the proposed framework provides a practical basis for cross-dataset transfer in heterogeneous HAR, while more effective multi-dataset pretraining strategies remain an important direction for future research.

\section{Discussion}
The results provide three main insights into channel-free HAR under heterogeneous sensor configurations. First, fusion design is a central issue under the channel-free constraint. Once the model no longer assumes a fixed channel order or composition, the fusion strategy directly determines the inductive bias of the model. In this study, the LF-based design achieved a favorable balance between flexibility and robustness, especially under channel perturbations such as missing channels and channel shuffling. This suggests that preserving channel-wise independence before aggregation is an effective design principle for robust channel-free HAR.

Second, metadata conditioning was beneficial but not universally reliable. Incorporating metadata improved the LF-based model in the main comparison, robustness analysis, and transfer learning experiments, suggesting that metadata can compensate for structural information lost in channel-independent processing. However, the perturbation-intensity analysis also showed that this benefit depends on correct signal--metadata alignment. When metadata were corrupted, the advantage gradually diminished, but it did not collapse catastrophically: the bounded identity-perturbation form of CBN in Eq.~\eqref{eq:cbn} ensures that, in the limit of uninformative metadata, the layer degenerates to plain BatchNorm, which provides a graceful fallback. Still, metadata should be treated as useful side information rather than a fully reliable source of structure, and ensuring metadata quality or designing conditioning mechanisms that are even more tolerant of misalignment are worthwhile engineering considerations for deployment.

Third, the transfer learning results highlight an important implication of channel-free modeling. Because the proposed framework does not rely on a fixed input channel structure, it can be transferred across datasets with different sensor configurations. This is particularly relevant in light of recent HAR studies that seek to exploit multiple datasets through domain generalization and data-mixtured pretraining. The positive single-source transfer results support this advantage. However, simple multitask pretraining was not sufficient to fully exploit it, suggesting that label-space alignment and more principled multi-dataset pretraining remain important future challenges, as also implied by recent efforts on mixed-dataset HAR pretraining \cite{Zhao2022_DRP, ban2025, Miao2024, Qiu2025}.

Several limitations should also be noted, and we organize them along three axes that we consider the most important future directions.
First, on \textbf{metadata integration}, the current CBN-based conditioning does not exploit all metadata types equally well (Section~\ref{sec:sensitivity}), and sensor-type information did not always contribute positively. A more expressive conditioning mechanism, such as attention-based metadata fusion or contrastive metadata--signal alignment, is a promising direction for future work.
Second, on \textbf{generalization to truly unseen sensor modalities}, although the proposed framework supports unseen channel counts and permutations out of the box, sensor types that are absent from training (e.g., EMG, barometric pressure) still require initializing and training new metadata embeddings. Extending the framework so that new modalities can be incorporated in a zero-shot or few-shot manner, for instance through natural-language descriptions of sensor semantics, would further strengthen the channel-free property.
Third, on \textbf{real-world IoT deployment}, the experiments here focused on classification benchmarks and controlled perturbation settings, and did not cover issues such as sensor drift, clock synchronization errors, and packet loss that frequently arise in wearable and ambient IoT systems. Evaluating and, if necessary, extending the proposed design under these operational stresses is an important direction for bringing channel-free HAR closer to practice.

\section{Conclusion}
In this study, we investigated channel-free HAR under heterogeneous sensor configurations and showed that this setting makes fusion design a central challenge rather than a simple input-format issue. To address this problem, we proposed a channel-free HAR framework based on channel-wise encoding, metadata-conditioned late fusion, and joint supervision of channel-level and fused predictions.

The experimental results demonstrated that the proposed design achieves a favorable balance between robustness and performance. In particular, it remained stable under channel perturbations, benefited from metadata when signal--metadata alignment was preserved, and enabled transfer across datasets with different sensor configurations. These findings suggest that inductive-bias-aware channel-free modeling is a promising direction for robust and transferable HAR in heterogeneous IoT environments.

\bibliographystyle{IEEEtran}
\bibliography{mybib}

@article{Kwapisz2011,
 author = {Kwapisz, Jennifer R. and others},
 title = {Activity Recognition Using Cell Phone Accelerometers},
 journal = {SIGKDD Explor. Newsl.},
 issue_date = {December 2010},
 volume = {12},
 number = {2},
 year = {2011},
 issn = {1931-0145},
 pages = {74--82},
 doi = {10.1145/1964897.1964918},
}

@Article{Shoaib2016,
AUTHOR = {Shoaib, Muhammad and others},
TITLE = {Complex Human Activity Recognition Using Smartphone and Wrist-Worn Motion Sensors},
JOURNAL = {Sensors},
VOLUME = {16},
YEAR = {2016},
NUMBER = {4},
ARTICLE-NUMBER = {426},
}

@Article{Robert2019,
AUTHOR = { R.-A. Voicu and others},
TITLE = {Human Physical Activity Recognition Using Smartphone Sensors},
JOURNAL = {Sensors},
VOLUME = {19},
YEAR = {2019},
NUMBER = {3},
ARTICLE-NUMBER = {458},
}

@inproceedings{Kawaguchi2011,
  title={HASC Challenge: Gathering Large Scale Human Activity Corpus for the Real-World Activity Understandings},
  author={N. Kawaguchi and others},
  booktitle={In Proc. of the 2nd Augmented Human International Conference},
  year={2011},
	month={Mar.},
	location={Tokyo, Japan}
}

@Article{Daniela2017,
    AUTHOR = {D. Micucci and others},
    TITLE = {UniMiB SHAR: A Dataset for Human Activity Recognition Using Acceleration Data from Smartphones},
    JOURNAL = {Applied Sciences},
    VOLUME = {7},
    YEAR = {2017},
    NUMBER = {10},
}

@INPROCEEDINGS{Stisen2015-ai,
  title     = "Smart devices are different: Assessing and {MitigatingMobile} sensing heterogeneities for activity recognition",
  author    = "Stisen, Allan and others",
  booktitle = "Proceedings of the 13th ACM Conference on Embedded Networked Sensor Systems",
  publisher = "ACM",
  address   = "New York, NY, USA",
  month     =  nov,
  year      =  2015
}

@INCOLLECTION{Banos2014-uc,
  title     = "{MHealthDroid}: A novel framework for agile development of mobile
               health applications",
  author    = "Banos, Oresti and others",
  booktitle = "Ambient Assisted Living and Daily Activities",
  publisher = "Springer International Publishing",
  address   = "Cham",
  pages     = "91--98",
  series    = "Lecture Notes in Computer Science",
  year      =  2014
}

@inproceedings{Reiss2012a,
 author = {A. Reiss and D. Stricker},
 title = {Creating and Benchmarking a New Dataset for Physical Activity Monitoring},
 booktitle = {In Proc. of the 5th International Conference on PErvasive Technologies Related to Assistive Environments (PETRA)},
 year = {2012},
 location = {Heraklion, Crete, Greece},
 pages = {40:1--40:8},
 doi = {10.1145/2413097.2413148},
}

@INPROCEEDINGS{Zhao2022_DRP,
  author={Zhao, Zhongkai and Hasegawa, Tatsuhito},
  booktitle={2022 International Conference on Machine Learning and Cybernetics (ICMLC)}, 
  title={Domain-Robust Pre-Training Method for the Sensor-Based Human Activity Recognition}, 
  year={2022},
  volume={},
  number={},
  pages={67-71},
}

@misc{bommasani2022opportunities,
      title={On the Opportunities and Risks of Foundation Models}, 
      author={Rishi Bommasani and others},
      year={2022},
      eprint={2108.07258},
      archivePrefix={arXiv},
      primaryClass={cs.LG}
}

@ARTICLE{Zhao2022,
  author={Zhao, Zhongkai and others},
  journal={IEEE Access}, 
  title={A Comparative Study: Toward an Effective Convolutional Neural Network Architecture for Sensor-Based Human Activity Recognition}, 
  year={2022},
  volume={10},
  number={},
  pages={20547-20558},
  doi={10.1109/ACCESS.2022.3152530}
}

@ARTICLE{Zhao2023-kt,
  title     = "Attention‐based sensor fusion for emotion recognition from human motion by combining convolutional neural network and weighted kernel support vector machine and using inertial measurement unit signals",
  author    = "Zhao, Yan and others",
  journal   = "IET Signal Proc.",
  volume    =  17,
  number    =  4,
  month     =  apr,
  year      =  2023,
}

@ARTICLE{Gholamrezaii2021-da,
  title    = "A time-efficient convolutional neural network model in human activity recognition",
  author   = "Gholamrezaii, Marjan and AlModarresi, S M T",
  journal  = "Multimedia Tools and Applications",
  volume   =  80,
  number   =  13,
  pages    = "19361--19376",
  year     =  2021
}

@ARTICLE{Yang2018-co,
  title     = "{DFTerNet}: Towards 2-bit Dynamic Fusion Networks for Accurate Human Activity Recognition",
  author    = "Yang, Zhan and others",
  journal   = "IEEE Access",
  publisher = "IEEE",
  volume    =  6,
  pages     = "56750--56764",
  year      =  2018
}

@INPROCEEDINGS{Munzner2017-uw,
  title     = "{CNN-based} sensor fusion techniques for multimodal human activity recognition",
  booktitle = "Proceedings of the 2017 {ACM} International Symposium on Wearable Computers",
  author    = "M{\"u}nzner, Sebastian and others",
  pages     = "158--165",
  series    = "ISWC '17",
  month     =  sep,
  year      =  2017,
}

@ARTICLE{Baloch2022-sl,
  title     = "{CNN-LSTM-Based} Late Sensor Fusion for Human Activity Recognition in Big Data Networks",
  author    = "Baloch, Zartasha and Shaikh, Faisal Karim and Unar, Mukhtiar Ali",
  journal   = "Wireless Communications and Mobile Computing",
  volume    =  2022,
  month     =  aug,
  year      =  2022,
  language  = "en"
}

@ARTICLE{Wan2020-iz,
  title    = "Deep Learning Models for Real-time Human Activity Recognition
              with Smartphones",
  author   = "Wan, Shaohua and and others",
  journal  = "Mobile Networks and Applications",
  volume   =  25,
  number   =  2,
  pages    = "743--755",
  month    =  apr,
  year     =  2020
}

@ARTICLE{Dua2021-dz,
  title    = "Multi-input {CNN-GRU} based human activity recognition using
              wearable sensors",
  author   = "Dua, Nidhi and others",
  journal  = "Computing",
  volume   =  103,
  number   =  7,
  pages    = "1461--1478",
  month    =  jul,
  year     =  2021
}

@Inproceedings{Simonyan2014,
Author = {K. Simonyan and A. Zisserman},
Title = {Very Deep Convolutional Networks for Large-Scale Image Recognition},
booktitle = 	 {Proc. of the International Conference on Learning Representations},
pages = 	 {1--14},
year = 	 {2015},
month = 	 {May},
location = {San Diego, CA, USA},
}

@ARTICLE{Anguita2013-qb,
  title   = "A public domain dataset for human activity recognition using smartphones",
  author  = "Anguita, D and others",
  journal = "European Symposium on Artificial Neural Networks (ESANN)",
  pages   = "437--442",
  year    =  2013
}

@ARTICLE{Altun2010-kp,
  title     = "Comparative study on classifying human activities with miniature inertial and magnetic sensors",
  author    = "Altun, Kerem and others",
  journal   = "Pattern Recognit.",
  publisher = "Elsevier BV",
  volume    =  43,
  number    =  10,
  pages     = "3605--3620",
  month     =  oct,
  year      =  2010,
}

@InProceedings{Dong2019,
author="Dong, Mingtao and others",
title="HAR-Net: Fusing Deep Representation and Hand-Crafted Features for Human Activity Recognition",
booktitle="Signal and Information Processing, Networking and Computers",
year="2019",
publisher="Springer Singapore",
address="Singapore",
pages="32--40",
}

@Article{Tuncer2020,
author={Tuncer, Turker and others},
title={Ensemble residual network-based gender and activity recognition method with signals},
journal={The Journal of Supercomputing},
year={2020},
month={Mar},
day={01},
volume={76},
number={3},
pages={2119-2138},
}

@ARTICLE{Ronald2021,
  author={Ronald, Mutegeki and others},
  journal={IEEE Access}, 
  title={iSPLInception: An Inception-ResNet Deep Learning Architecture for Human Activity Recognition}, 
  year={2021},
  volume={9},
  number={},
  pages={68985-69001},
}

@ARTICLE{Miah2024,
  author={Miah, Abu Saleh Musa and others},
  journal={IEEE Access}, 
  title={Sensor-Based Human Activity Recognition Based on Multi-Stream Time-Varying Features With ECA-Net Dimensionality Reduction}, 
  year={2024},
  volume={12},
  number={},
}

@ARTICLE{Li2025,
  author={Li, Shuangjian and others},
  journal={IEEE Internet of Things Journal}, 
  title={P2LHAP: Wearable-Sensor-Based Human Activity Recognition, Segmentation, and Forecast Through Patch-to-Label Seq2Seq Transformer}, 
  year={2025},
  volume={12},
  number={6},
  pages={6818-6830},
}

@article{YANG2025108314,
title = {Enhancing human activity recognition with TB-ConvAtt: A multi-dimensional attention framework},
journal = {Biomedical Signal Processing and Control},
volume = {110},
pages = {108314},
year = {2025},
author = {Hongmei Yang and others},
}

@article{Noori2020,
author = {Noori, Farzan Majeed and others},
title = {Human Activity Recognition from Multiple Sensors Data Using Multi-fusion Representations and CNNs},
journal = {ACM Trans. Multimedia Comput. Commun. Appl.},
year = {2020},
volume = {16},
number = {2},
articleno = {45},
numpages = {19},
}

@InProceedings{Kasnesis2019,
author="Kasnesis, Panagiotis and others",
title="PerceptionNet: A Deep Convolutional Neural Network for Late Sensor Fusion",
booktitle="Intelligent Systems and Applications",
year="2019",
publisher="Springer International Publishing",
address="Cham",
pages="101--119",
}

@InProceedings{Hasegawa2024,
author="Hasegawa, Tatsuhito",
title="Human Activity Recognition Model Capable of Handling Various Input Waveforms",
booktitle="Neural Information Processing",
year="2025",
publisher="Springer Nature Singapore",
address="Singapore",
pages="1--16",
}

@article{Qin2020,
author = {Qin, Xin and others},
title = {Cross-Dataset Activity Recognition via Adaptive Spatial-Temporal Transfer Learning},
year = {2020},
volume = {3},
number = {4},
journal = {Proc. ACM Interact. Mob. Wearable Ubiquitous Technol.},
articleno = {148},
numpages = {25},
}

@article{Hong2024,
author = {Hong, Zhiqing and others},
title = {CrossHAR: Generalizing Cross-dataset Human Activity Recognition via Hierarchical Self-Supervised Pretraining},
year = {2024},
address = {New York, NY, USA},
volume = {8},
number = {2},
journal = {Proc. ACM Interact. Mob. Wearable Ubiquitous Technol.},
articleno = {64},
numpages = {26},
}

@article{Miao2024,
author = {Miao, Shenghuan and Chen, Ling},
title = {GOAT: A Generalized Cross-Dataset Activity Recognition Framework with Natural Language Supervision},
year = {2024},
volume = {8},
number = {4},
journal = {Proc. ACM Interact. Mob. Wearable Ubiquitous Technol.},
articleno = {184},
numpages = {28},
}

@InProceedings{SAM2023,
    author    = {Kirillov, Alexander and others},
    title     = {Segment Anything},
    booktitle = {Proceedings of the IEEE/CVF International Conference on Computer Vision (ICCV)},
    month     = {October},
    year      = {2023},
    pages     = {4015-4026}
}

@InProceedings{GDINO2025,
author="Liu, Shilong and others",
title="Grounding DINO: Marrying DINO with Grounded Pre-training for Open-Set Object Detection",
booktitle="European Conference on Computer Vision (ECCV) 2024",
year="2025",
publisher="Springer Nature Switzerland",
address="Cham",
pages="38--55",
isbn="978-3-031-72970-6"
}

@misc{GPT2018,
  title        = {Improving Language Understanding by Generative Pre-Training},
  author       = {Radford, Alec and others},
  year         = {2018},
  howpublished = {OpenAI Technical Report},
  url          = {https://cdn.openai.com/research-covers/language-unsupervised/language_understanding_paper.pdf},
  note         = {Accessed: 2026-01-14}
}

@misc{ban2025,
      title={HAR-DoReMi: Optimizing Data Mixture for Self-Supervised Human Activity Recognition Across Heterogeneous IMU Datasets}, 
      author={Lulu Ban and others},
      year={2025},
      eprint={2503.13542},
      archivePrefix={arXiv},
      primaryClass={cs.LG},
      url={https://arxiv.org/abs/2503.13542}, 
}

@inproceedings{Inoue2015,
author = {Inoue, Sozo and others},
title = {Mobile activity recognition for a whole day: recognizing real nursing activities with big dataset},
year = {2015},
booktitle = {Proceedings of the 2015 ACM International Joint Conference on Pervasive and Ubiquitous Computing (UbiComp)},
pages = {1269–1280},
numpages = {12},
}

@inproceedings{Maekawa2016,
author = {Maekawa, Takuya and others},
title = {Toward practical factory activity recognition: unsupervised understanding of repetitive assembly work in a factory},
year = {2016},
booktitle = {Proceedings of the 2016 ACM International Joint Conference on Pervasive and Ubiquitous Computing (UbiComp)},
pages = {1088–1099},
numpages = {12},
}

@inproceedings{Li2020,
author = {Li, Xi and others},
title = {ActivityGAN: generative adversarial networks for data augmentation in sensor-based human activity recognition},
year = {2020},
booktitle = {Adjunct Proceedings of the 2020 ACM International Joint Conference on Pervasive and Ubiquitous Computing and Proceedings of the 2020 ACM International Symposium on Wearable Computers (UbiComp/ISWC)},
pages = {249–254},
numpages = {6},
}

@article{RASHID2019100944,
title = {Times-series data augmentation and deep learning for construction equipment activity recognition},
journal = {Advanced Engineering Informatics},
volume = {42},
pages = {100944},
year = {2019},
author = {Khandakar M. Rashid and Joseph Louis},
}

@ARTICLE{Hasegawa2021,
  author={Hasegawa, Tatsuhito},
  journal={IEEE Access}, 
  title={Octave Mix: Data Augmentation Using Frequency Decomposition for Activity Recognition}, 
  year={2021},
  volume={9},
  number={},
  pages={53679-53686},
}

@Article{Varshney2022,
author={Varshney, Neeraj and others},
title={Human activity recognition using deep transfer learning of cross position sensor based on vertical distribution of data},
journal={Multimedia Tools and Applications},
year={2022},
month={Jul},
day={01},
volume={81},
number={16},
pages={22307-22322},
}

@ARTICLE{Zhao2021,
  author={Zhao, Jiachen and others},
  journal={IEEE Transactions on Human-Machine Systems}, 
  title={Local Domain Adaptation for Cross-Domain Activity Recognition}, 
  year={2021},
  volume={51},
  number={1},
  pages={12-21},
}

@article{Qiu2025,
author = {Qiu, Minghui and others},
title = {Towards Customizable Foundation Models for Human Activity Recognition with Wearable Devices},
year = {2025},
volume = {9},
number = {3},
journal = {Proc. ACM Interact. Mob. Wearable Ubiquitous Technol.},
articleno = {122},
numpages = {29},
}

@inproceedings{
zhang2025sensorlm,
title={Sensor{LM}: Learning the Language of Wearable Sensors},
author={Yuwei Zhang and others},
booktitle={The Thirty-ninth Annual Conference on Neural Information Processing Systems (NeurIPS)},
year={2025},
}

@inproceedings{mo2024deepavfusion,
  title={Unveiling the Power of Audio-Visual Early Fusion Transformers with Dense Interactions through Masked Modeling},
  author={Mo, Shentong and Morgado, Pedro},
  booktitle={Proceedings of the IEEE/CVF Conf. on Computer Vision and Pattern Recognition (CVPR)},
  year={2024}
}

@article{Li2024AV,
author = {Li, Yidi and others},
title = {Audio--visual keyword transformer for unconstrained sentence-level keyword spotting},
journal = {CAAI Transactions on Intelligence Technology},
volume = {9},
number = {1},
pages = {142-152},
year = {2024}
}

@inproceedings{NIPS2017_f22e4747,
 author = {Zaheer, Manzil and others},
 booktitle = {Advances in Neural Information Processing Systems},
 editor = {I. Guyon and U. Von Luxburg and S. Bengio and H. Wallach and R. Fergus and S. Vishwanathan and R. Garnett},
 pages = {},
 publisher = {Curran Associates, Inc.},
 title = {Deep Sets},
 volume = {30},
 year = {2017}
}

@inproceedings{Harm2017CBN,
author = {de Vries, Harm and Strub, Florian and Mary, J\'{e}r\'{e}mie and Larochelle, Hugo and Pietquin, Olivier and Courville, Aaron},
title = {Modulating early visual processing by language},
year = {2017},
isbn = {9781510860964},
publisher = {Curran Associates Inc.},
address = {Red Hook, NY, USA},
booktitle = {Proceedings of the 31st International Conference on Neural Information Processing Systems},
pages = {6597–6607},
numpages = {11},
location = {Long Beach, California, USA},
series = {NIPS'17}
}

@InProceedings{Lea_2017_CVPR,
author = {Lea, Colin and Flynn, Michael D. and Vidal, Rene and Reiter, Austin and Hager, Gregory D.},
title = {Temporal Convolutional Networks for Action Segmentation and Detection},
booktitle = {Proceedings of the IEEE Conference on Computer Vision and Pattern Recognition (CVPR)},
month = {July},
year = {2017}
}

@InProceedings{Sandler_2018_CVPR,
author = {Sandler, Mark and Howard, Andrew and Zhu, Menglong and Zhmoginov, Andrey and Chen, Liang-Chieh},
title = {MobileNetV2: Inverted Residuals and Linear Bottlenecks},
booktitle = {Proceedings of the IEEE Conference on Computer Vision and Pattern Recognition (CVPR)},
month = {June},
year = {2018}
}

@InProceedings{pmlr-v97-tan19a,
  title = 	 {{E}fficient{N}et: Rethinking Model Scaling for Convolutional Neural Networks},
  author =       {Tan, Mingxing and Le, Quoc},
  booktitle = 	 {Proceedings of the 36th International Conference on Machine Learning},
  pages = 	 {6105--6114},
  year = 	 {2019},
  editor = 	 {Chaudhuri, Kamalika and Salakhutdinov, Ruslan},
  volume = 	 {97},
  series = 	 {Proceedings of Machine Learning Research},
  month = 	 {09--15 Jun},
  publisher =    {PMLR},
}

@inproceedings{takama2026dynamic,
  author    = {Hiromu Takama and Tatsuhito Hasegawa},
  title     = {Dynamic Negative Correlation Learning in Deep Ensemble Learning},
  booktitle = {Proceedings of the 14th IIAE International Conference on Industrial Application Engineering 2026},
  year      = {2026},
  publisher = {The Institute of Industrial Applications Engineers, Japan}
}

@inproceedings{Locatello2020SlotoAttn,
author = {Locatello, Francesco and Weissenborn, Dirk and Unterthiner, Thomas and Mahendran, Aravindh and Heigold, Georg and Uszkoreit, Jakob and Dosovitskiy, Alexey and Kipf, Thomas},
title = {Object-centric learning with slot attention},
year = {2020},
isbn = {9781713829546},
publisher = {Curran Associates Inc.},
address = {Red Hook, NY, USA},
booktitle = {Proceedings of the 34th International Conference on Neural Information Processing Systems},
articleno = {967},
numpages = {14},
location = {Vancouver, BC, Canada},
series = {NIPS '20}
}

@INPROCEEDINGS{He2016resnet,
  author={He, Kaiming and Zhang, Xiangyu and Ren, Shaoqing and Sun, Jian},
  booktitle={2016 IEEE Conference on Computer Vision and Pattern Recognition (CVPR)}, 
  title={Deep Residual Learning for Image Recognition}, 
  year={2016},
  volume={},
  number={},
  pages={770-778},
  doi={10.1109/CVPR.2016.90}}

\begin{IEEEbiography}[{\includegraphics[width=1in,height=1.25in,clip,keepaspectratio]{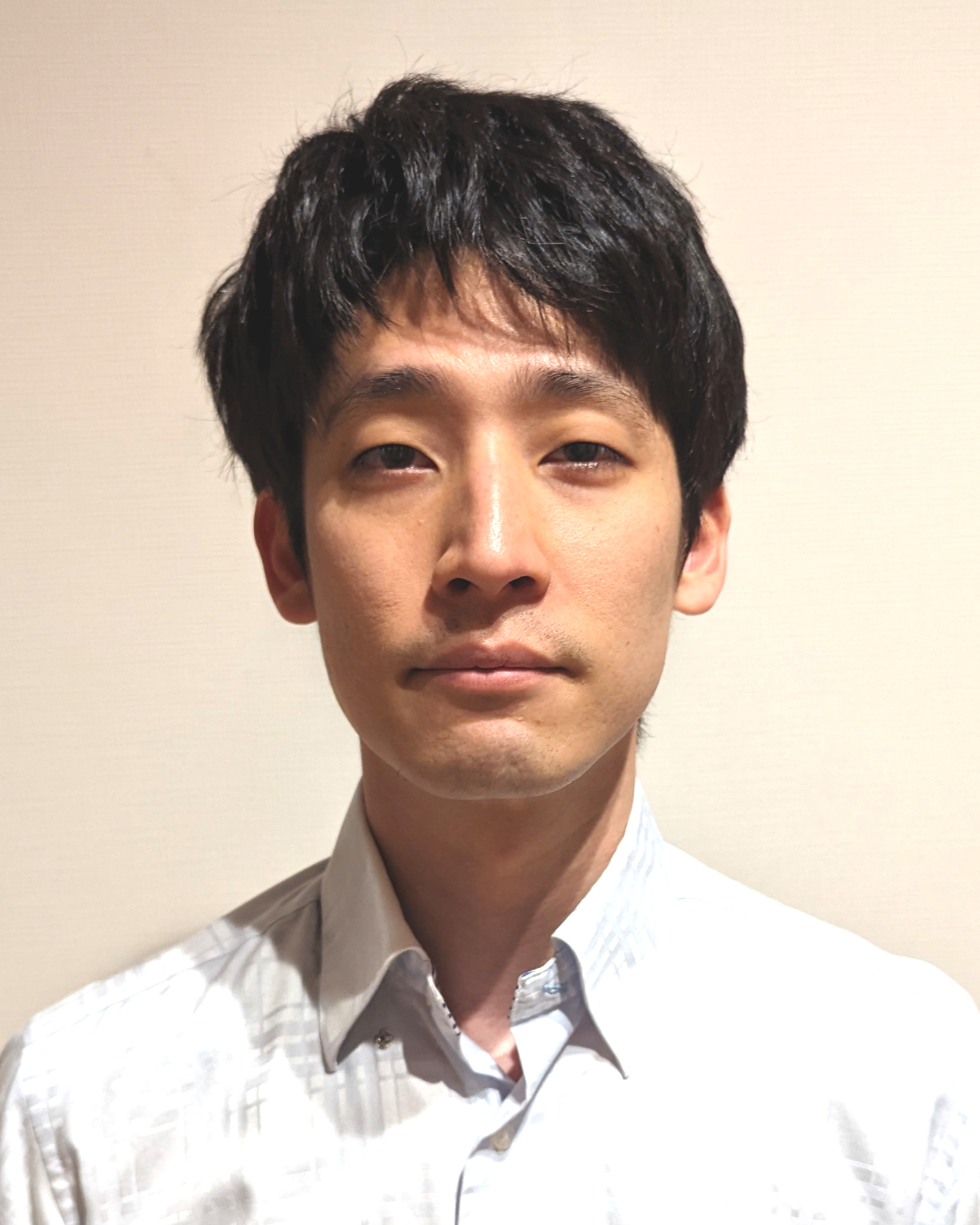}}]{Tatsuhito Hasegawa}
  (Member, IEEE) received the Ph.D. degree in engineering from Kanazawa University, Kanazawa, in 2015. From 2011 to 2013, he was a System Engineer with Fujitsu Hokuriku Systems Ltd. From 2014 to 2017, he was an Assistant with Tokyo Healthcare University. From 2017 to 2020, he was a Senior Lecturer with the Graduate School of Engineering, University of Fukui. He is currently an Associate Professor. His study interests include human activity recognition, deep learning, and intelligent learning support system. He is also a member of IPSJ and JSAI.
\end{IEEEbiography}

\vfill

\end{document}